\documentclass[journal]{IEEEtran}

\usepackage{cite}

\ifCLASSINFOpdf
 \usepackage[pdftex]{graphicx}
 \graphicspath{figures}
 \DeclareGraphicsExtensions{.pdf,.jpeg,.png}
\else
\fi

\usepackage{tikz}
\usepackage{amsmath,amsfonts,amssymb}
\usepackage{colortbl}
\usepackage{comment}
\usepackage{multirow}
\usepackage{url}
\usepackage{algorithm}
\usepackage{algorithmicx}
\usepackage{textcomp}
\usepackage{hyperref}
\usepackage[caption=false]{subfig}
\usepackage{booktabs}
\usepackage{tabularx}

\usepackage{stfloats}
\usepackage{comment}

\begin{document}

\title{Weakly-Supervised Surface Crack Segmentation by Generating Pseudo-Labels using Localization with a Classifier and Thresholding}

\author{Jacob K\"onig,
        Mark David Jenkins,
        Mike Mannion,
        Peter Barrie,
        and Gordon Morison
\thanks{The authors are with the School of Computing, Engineering and Built Environment, Glasgow Caledonian University, G4 0BA Glasgow, Scotland (Email: jacob.koenig@gcu.ac.uk).}}

\maketitle

\begin{abstract}
Surface cracks are a common sight on public infrastructure nowadays. Recent work has been addressing this problem by supporting structural maintenance measures using machine learning methods. Those methods are used to segment surface cracks from their background, making them easier to localize. However, a common issue is that to create a well-functioning algorithm, the training data needs to have detailed annotations of pixels that belong to cracks. Our work proposes a weakly supervised approach that leverages a CNN classifier in a novel way to create surface crack pseudo labels. First, we use the classifier to create a rough crack localization map by using its class activation maps and a patch based classification approach and fuse this with a thresholding based approach to segment the mostly darker crack pixels. The classifier assists in suppressing noise from the background regions, which commonly are incorrectly highlighted as cracks by standard thresholding methods. Then, the pseudo labels can be used in an end-to-end approach when training a standard CNN for surface crack segmentation. Our method is shown to yield sufficiently accurate pseudo labels. Those labels, incorporated into segmentation CNN training using multiple recent crack segmentation architectures, achieve comparable performance to fully supervised methods on four popular crack segmentation datasets.
\end{abstract}

\section{Introduction}
Cracks are a type of surface defect and present an indication of structural stress in artifacts within the built environment, often indicating failure. If left untreated, cracks can grow in size and may pose a risk to those structures. This is becoming ever more important as roads, buildings, and bridges are used beyond their lifespan and lack the required maintenance \cite{2020COVID19Impacts, homer2018ThousandsMiles}. Much research within this space has focused on the classification and segmentation of surface cracks, however, the majority of those works published make use of supervised learning \cite{inoue2019DeploymentConsciousa, liu2019DeepCrackDeepa,konig2021OptimizedDeep}.  A prevalent issue with supervised learning is that the manual creation of annotations can take up a large amount of time, especially when annotations for segmentation are performed on a pixel level \cite{inoue2020CrackDetection}. 
Recent research has found a way around this problem in the form of using semi or weakly-supervised learning. Using those methods, the required effort to label images has been reduced significantly as images only using classification labels, images containing rough annotations, or datasets with a small labeled subset can be used to train methods to perform segmentation. 
Early algorithms in this space did not include automated learning or used handcrafted features to segment cracks. Examples include basic mathematical morphology \cite{tanaka1998CrackDetection}, thresholding \cite{oliveira2009AutomaticRoad}, edge detection \cite{zhao2010ImprovementCanny} and Wavelet decomposition \cite{wang2007WaveletBasedPavement} to segment surface cracks from their background. The issue with some of those more traditional methods is that whilst they perform well in segmenting cracks, they also segment background regions. This is because a majority of cracks are darker than their surroundings, leading to darker, non-crack spots incorrectly being segmented.
Learning-based methods, especially deep learning methods, have followed those traditional approaches and shown much better performance and adaptability to crack features. This started with some works performing a pixel-wise classification \cite{fan2018AutomaticPavementa, inoue2019DeploymentConsciousa} and later switched to using purpose-built segmentation architectures \cite{jenkins2018DeepConvolutionala, liu2019DeepCrackDeepa, konig2021OptimizedDeep}. However, a common drawback of those deep learning methods is that they require large amounts of labeled data to train, before subsequently being used to perform inference on novel data. 

To overcome the effort of labeling data to train segmentation algorithms, we propose a novel method that leverages the strong performance of a basic crack classifier convolutional neural network (CNN), and only needs to be trained using classification labels to consequently create segmentation maps. We combine this classifier with a patch-based approach and some traditional techniques to create a weakly-supervised crack segmentation method. We aim to mitigate the problem of more traditional methods: 1) that background regions are segmented which ``stain" the segmentation map and 2) there is a high labeling effort inherent in fully-supervised methods.
Our method focuses on simplicity and can be easily implemented if a pre-trained crack classifier is available. To support this and further research, we also provide the code implementation for our work\footnote{Code \textbf{will be made} available at: \url{https://github.com/jacobkoenig/WeaklySupervisedCrackSeg}}.
For our proposed method we make use of two assumptions: 1) that cracks differ in their contrast and brightness from the surrounding areas and 2), the crack classification model is trained well enough to detect a majority of surface cracks on a given task/dataset.
We acknowledge that these assumptions may not hold for all use-cases or surface cracks, but this seems to be a common problem as even trained, fully-supervised segmentation algorithms have difficulty segmenting brighter cracks \cite{konig2021OptimizedDeep}.
Therefore, the contributions of this work can be summarized as follows:
\begin{enumerate}
    \item Weakly-supervised crack segmentation is achieved by only using classification labels to train a classifier.
    \item The classifier is leveraged in two ways to generate coarse localization maps of surface cracks. This is achieved by using a patch-based classification approach to classify individual regions of an image and merging that with the class activation map of the classifier.
    \item The coarse localization maps are then merged with a segmentation map that uses a novel patch-based thresholding technique to separate the darker pixels, which belong to cracks, from background pixels. For this, filtering for noise reduction and morphology for increasing crack connectivity are used before resulting in a pseudo label segmentation map.
    \item Those pseudo labels are then incorporated into end-to-end training for a standard semantic segmentation algorithm that generates segmentation maps for cracks. We show that our weakly-supervised method achieves competitive results on multiple datasets when compared against fully-supervised end-to-end training.
\end{enumerate}

\section{Related Work}
In the last few years, crack segmentation algorithms were mostly based on CNN and fully-supervised learning. Fan \textit{et al.} \cite{fan2018AutomaticPavementa} propose a deep learning-based method, which classifies whether a crack is in the center pixel of a patch and then applies this algorithm for every pixel in an image. In \cite{inoue2019DeploymentConsciousa}, a similar approach is proposed, however, they also use their method to predict the orientation of cracks. Other popular crack segmentation architectures often follow an encoder-decoder-like structure for fully-supervised segmentation. In \cite{liu2019DeepCrackDeepa}, the encoder stage of this CNN is connected to several interim upsampling stages which, when combined and merged create an output segmentation map. The works in \cite{jenkins2018DeepConvolutionala, konig2019ConvolutionalNeurala} follow a patch-based encoder-decoder crack segmentation approach based on an U-Net architecture, which was initially proposed to segment biomedical data \cite{ronneberger2015UNetConvolutional}. The work in \cite{konig2021OptimizedDeep} uses a pretrained encoder network and studies the optimal composition of the decoder part to achieve the highest performance on surface crack segmentation. 

Recently, some works have also addressed the issue of semi and weakly-supervised semantic crack segmentation. 
The semi-supervised crack segmentation approaches in \cite{li2020SemiSupervisedSemantic, shim2020MultiscaleAdversarial} use adversarial learning and datasets where only a subset has been labeled. They propose to train a segmentation network in combination with a discriminator network; After training, the output of the discriminator is used as a supervised pseudo-label to train the segmentation algorithm on the unlabeled data. 
Inoue and Nagayoshi in \cite{inoue2020CrackDetection} propose a weakly-supervised approach, in which the output standard segmentation network architecture multiplied with a brightness-normalized image can create good segmentation results, even if the crack labels have been annotated much thicker. They exploit the fact that in a brightness-normalized image, darker parts (cracks) have higher values, which then limit the output of the segmentation network, even though it predicts cracks much thicker than they actually are. The weakly-supervised work in \cite{dong2020PatchBasedWeakly} uses a trained classifier to determine patch regions of images that belong to cracks. This is followed by observing the gradient activation maps and feeding them through a conditional random field to create pseudo-segmentation maps. Those pseudo-annotations are then used to train a fully-supervised segmentation network. This approach is most closely related to ours, however, we use an entirely different way to generate pseudo labels. Combining traditional with recent approaches, a multi-step weakly-supervised approach is proposed in \cite{fan2019RoadCrack}. Here a classifying network is used as the first step, separating images with cracks from ones without. On images with cracks, filtering is performed to remove noise and then the authors propose an adaptive thresholding approach to separate darker, crack pixels from the background.

\section{Method}
\begin{figure}[!tb]
    \centering
    \includegraphics[width=\linewidth,]{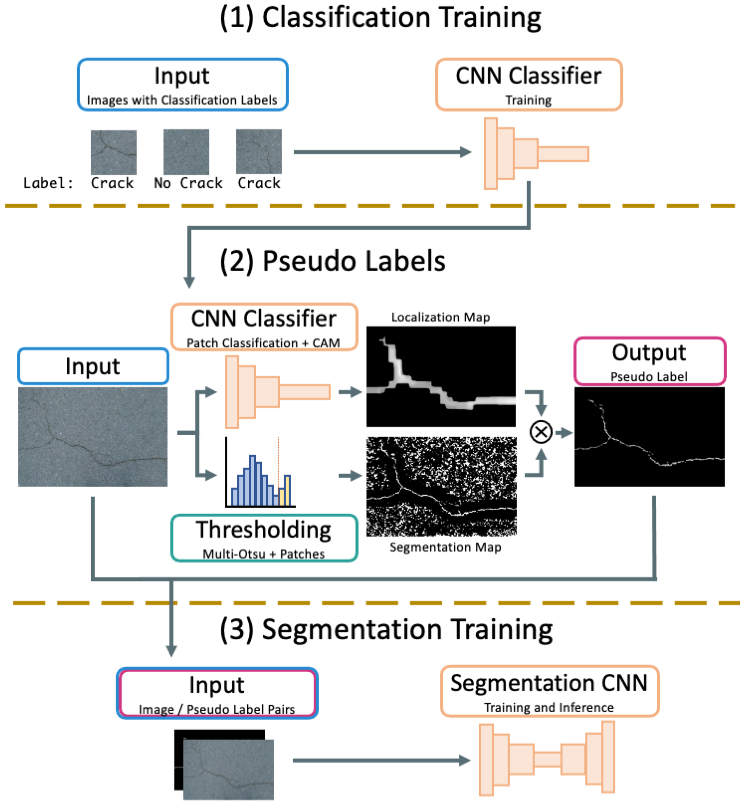}
    \caption{Overview of our proposed, weakly-supervised approach to create accurate segmentation maps when only classification labels are available.}
    \label{fig:method}
\end{figure}
Our method follows a multi-step approach in which solely having classification labels is sufficient to train an end-to-end segmentation CNN. It is based on using a trained CNN classifier, which is trained on at least one class belonging to cracks to infer pseudo-segmentation labels on the training set. Those pseudo-segmentation labels are then used to train a standard encoder-decoder CNN architecture for crack segmentation. 
This is a weakly-supervised method, as the label quality for training this classifier is significantly less than pixel-accurate segmentation labels. A general overview of our method is shown in \autoref{fig:method}.

\subsection{Classifier Training}
Since the approach in this work is using weakly-supervised learning, it is sufficient to only have a labeled image dataset with classification labels. Here, one class of those labels needs to belong to cracks. For the first step in our weakly-supervised approach, this image dataset will be used to train a CNN classifier. Popular choices of classification CNNs are the family of residual networks (ResNet 50 - 152) \cite{he2016DeepResidual} or EfficientNet \cite{tan2019EfficientNetRethinking} based networks. Once a classifier has been trained it can be used in the next step of our proposed approach, to generate pseudo-labels. It is to note that the choice of classifier should yield sufficient gradient-activation maps using methods such as GradCam \cite{selvaraju2017GradCamVisual} or GradCam++ \cite{chattopadhay2018GradCamGeneralized} as this is a key component in this approach.

\subsection{Coarse Crack Localization}
The classifier is then used to generate a coarse localization map of a crack by extracting two distinct kinds of information. First, small overlapping patch regions of the image to be pseudo-labeled are passed through the classifier, to find regions belonging to cracks. Second, the gradient activation maps of the last convolutional layer, captured when the full image is passed through, are converted to confidence scores. This creates two region maps, both of which are merged to generate the localization map.

For localization through patch classification, the image is split into overlapping patches that are sufficiently small so that they can still be processed by the classifier but give enough information on their surrounding area so that the classifier can determine whether a crack is present or not.
To increase the crack localization in this step, we set the patch extraction stride to be less than the patch size, which leads to better error suppression as the results of overlapping regions can be averaged. If a patch lies in a border region, we use mirror padding to expand the size. The output of each of those overlapping patches is then averaged, which leads to a localization map that is smaller than the input image. This first localization map is then upsampled to the original image size. However, a drawback of this step is that small image regions for classification can lead to some noise, as very dark patches can be classified as cracks for example. 

To mitigate this, we also use a localization through gradient activation approach. For this, we pass the full input image through the classifier and observe and upsample the activation map at the last convolutional layer using the GradCam++ method \cite{chattopadhay2018GradCamGeneralized}.

Both of those separate localization maps are then merged through averaging, and we retain only regions where there is a confidence score of greater than 50\% that there is a crack. Additionally, we also perform a morphological erosion operation on the merged map, as it can be assumed that cracks are located close to the centerline of highlighted regions.

\begin{figure}[!tb]
\centering
\captionsetup[subfigure]{labelformat=empty}
\subfloat[Input Image]{\includegraphics[width=0.24\linewidth]{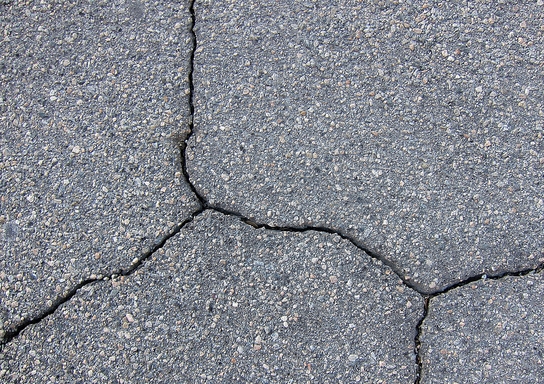}}\vspace{-0.05cm}
\subfloat[Patch Classification]{\includegraphics[width=0.24\linewidth]{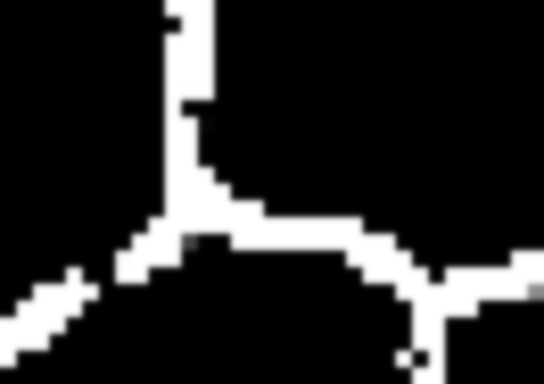}}\vspace{-0.05cm} 
\subfloat[Class Activations]{\includegraphics[width=0.24\linewidth]{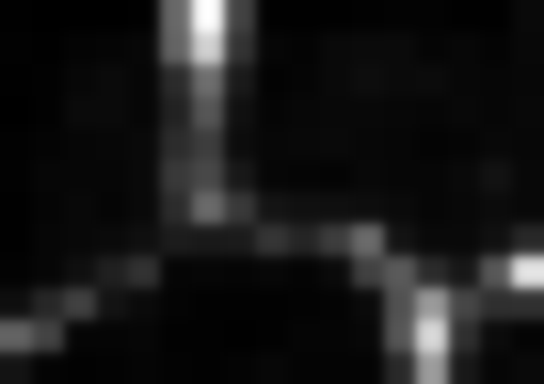}}\vspace{-0.08cm}
\subfloat[Localization Map]{\includegraphics[width=0.24\linewidth]{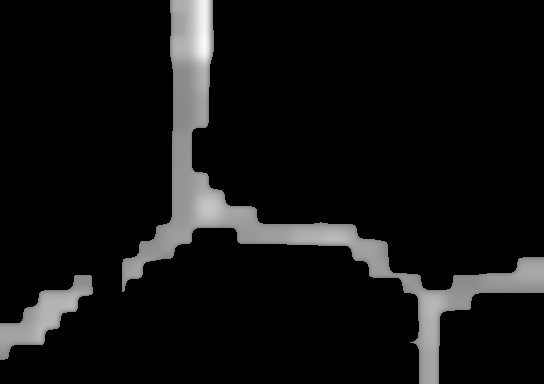}}\vspace{-0.05cm} \\
    \caption{Interim outputs of the steps to create the coarse localization map using the classifier.}
    \label{fig:localisation}
\end{figure}

We use the Lanczos method \cite{turkowski1990Filterscommon} for upsampling, as we have found that method to be better for preserving border regions over other methods.
A simple output of the sub-steps in our localization approach in \autoref{fig:localisation}.

\subsection{Crack Segmentation through Thresholding}
We then make use of the assumption that cracks stand out from the surrounding area, that a majority of them are of a darker color. We found that the normalization approach in \cite{inoue2020CrackDetection} can lead to adequate results, but it fails when the image is unevenly illuminated, since some actual regions of a crack may be as bright as regions that surround a crack in another part of the image.
We therefore propose to separate the actual pixels belonging to a crack following a sequential approach that makes use of patches of the image to be segmented. First, we pass the image through a bilateral filter \cite{tomasi1998BilateralFiltering}, similar to \cite{fan2019RoadCrack}, which removes some of the noise but provides good edge retention capabilities. Filtered pixels are generated by using a weighted average of neighboring pixels. The result of applying a bilateral filter $BF[\cdot]$ to an image $I$ is defined as:

\begin{equation}
    BF[I]_{(i,j)} = \sum_{x=k-d}^{x+d} \sum_{y=l-d}^{l+d} w(i,j,x,y) I_{(x,y)} 
\end{equation}
where
\begin{equation}
    w(i,j,x,y) = s(i,j,x,y) * r(i,j,x,y),
\end{equation}
\begin{equation}
    s(i,j,x,y) = \exp{-\frac{(i-x)^2+(j-y)^2}{2\sigma^2_s}},  
\end{equation}
\begin{equation}
    r(i,j,x,y) = \exp{-\frac{||I_{i,j} - I_{x,y}||^2}{2\sigma^2_r}} 
\end{equation}

Here, the subscripts $(i,j)$ and $(x,y)$ denote spatial pixel positions in the filtered or original image. The weighting coefficient $w(\cdot)$ is made up of the product of the spatial distance kernel $s(\cdot)$ and the range kernel $r(\cdot)$. The distance kernel takes into account the similarity between pixels at different spatial distances, whereas the range kernel uses the color intensity similarity. These kernels are modified by two hyperparameters $\sigma_s$ and $\sigma_r$. An additional hyperparameter $d$, determines the pixel neighborhood range that is used during filtering. 

\begin{figure}[!t]
\centering
\captionsetup[subfigure]{labelformat=empty}
\subfloat[Input Image]{\includegraphics[width=0.24\linewidth]{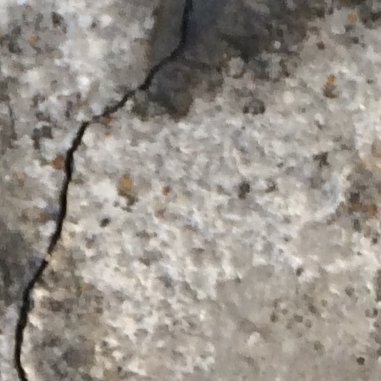}}\vspace{-0.04cm}
\subfloat[Two-Class Otsu \\(Full Image)]{\includegraphics[width=0.24\linewidth]{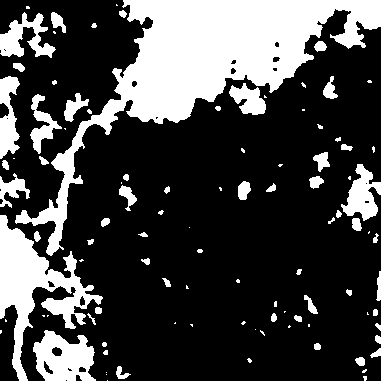}}\vspace{-0.04cm}
\subfloat[Multi Otsu \\(Full Image)]{\includegraphics[width=0.24\linewidth]{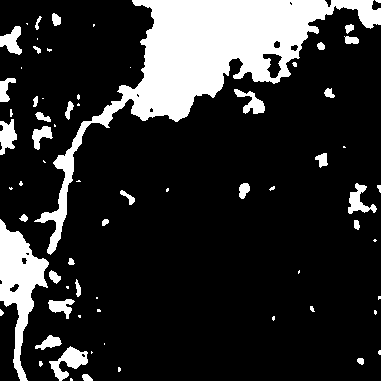}}\vspace{-0.04cm} 
\subfloat[Niblack \\(Full Image)]{\includegraphics[width=0.24\linewidth]{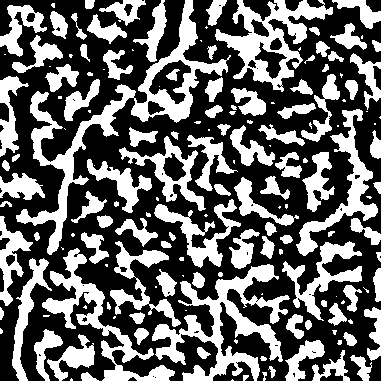}}\vspace{-0.04cm} \\
\subfloat[Sauvola \\ (Full Image)]{\includegraphics[width=0.24\linewidth]{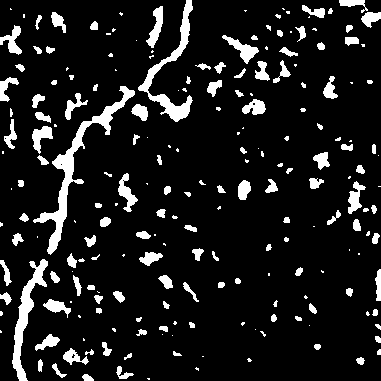}}\vspace{-0.04cm}
\subfloat[Two-Class Otsu \\ (Patches-Ours)]{\includegraphics[width=0.24\linewidth]{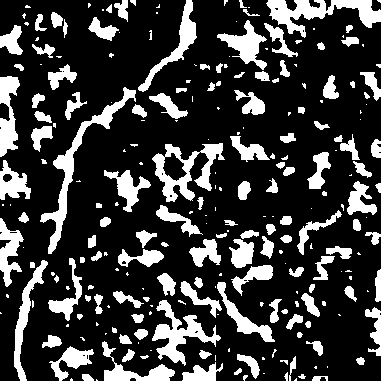}}\vspace{-0.04cm}
\subfloat[Multi Otsu \\ (Patches-Ours)]{\includegraphics[width=0.24\linewidth]{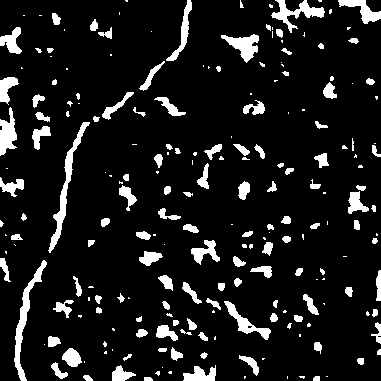}}\vspace{-0.04cm}
\subfloat[Ground Truth]{\includegraphics[width=0.24\linewidth]{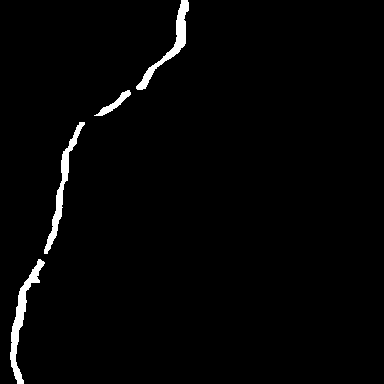}}\vspace{-0.04cm} \\
    \caption{Samples of performing different thresholding methods on a crack-image. Multi Otsu uses three classes. The Niblack and Sauvola methods use a window size of 33. Patches-Ours: Our local patch approach in which each patch is thresholded with a window size of 32 and a stride of 8 followed by combining the outputs.}
    \label{fig:thresholding_options}
\end{figure}
\begin{figure}[!t]
\centering
\captionsetup[subfigure]{labelformat=empty}
\subfloat[Input Image]{\includegraphics[width=0.24\linewidth]{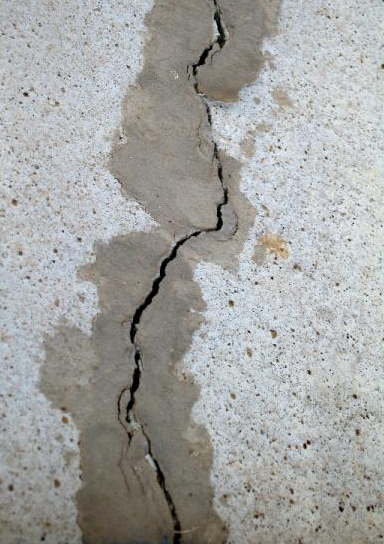}}\vspace{-0.04cm}
\subfloat[Localization Map]{\includegraphics[width=0.24\linewidth]{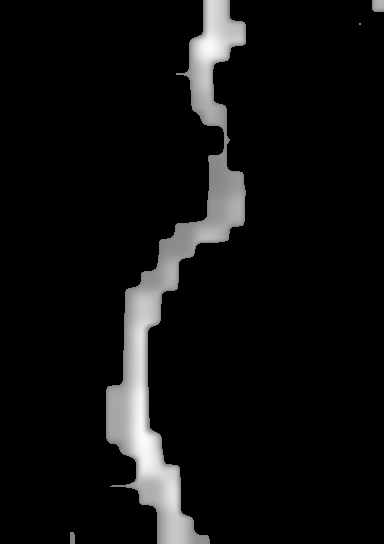}}\vspace{-0.04cm}
\subfloat[Segmentation Map]{\includegraphics[width=0.24\linewidth]{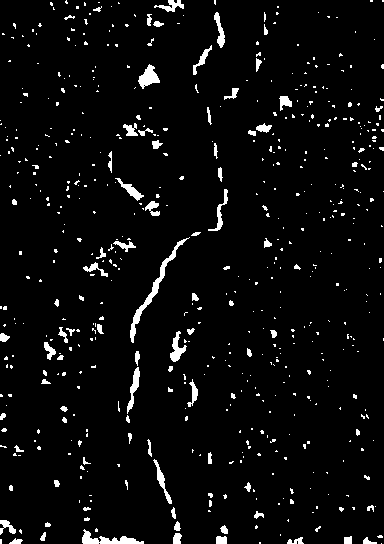}}\vspace{-0.04cm} 
\subfloat[Output]{\includegraphics[width=0.24\linewidth]{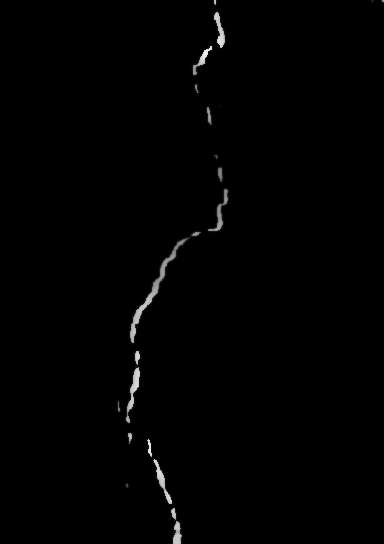}}\vspace{-0.04cm} \\
    \caption{Example output that has been generated by merging the localization map and the thresholded segmentation. }
    \label{fig:thresholding}
\end{figure}

Using the same patch size as the classifier, we then binarize the patch and obtain the crack pixels by using thresholding. 
Thresholding can be used to binarize images and can be set manually or using automatic methods such as Otsu. The basic (two-class) Otsu method automatically sets a threshold by maximizing the variance between classes of the image histogram.
We found that standard thresholding using a manual threshold, or using this basic Otsu thresholding did not lead to appropriate results as the threshold is different per patch and when set using the basic Otsu method it is usually too low. Instead, we make use of local, multi-Otsu thresholding \cite{otsu1979ThresholdSelection}. We posit that the pixel intensity of a normalized patch can be split into three categories: crack pixels (very dark), pixels that surround a crack (very light) and background regions (grey). Therefore a three-class multi-Otsu method seems appropriate. Following \cite{otsu1979ThresholdSelection}, using a grey level image with pixels intensity values $[1, \cdots, L]$, two thresholds $1 \leq k_1 \leq k_2 \leq L$ are utilised to separate the three classes: $C_1$ with $[1, \cdots, k_1]$, $C_2$ with $[k_1+1, \cdots, k_2]$ and $C_3$ with $[k_2+1, \cdots, L]$ .
The multi-Otsu method maximises the between class variance $\sigma^2_B$:
\begin{equation}
    \sigma^2_B = P_1(\mu_1 - \mu)^2 + P_2(\mu_2 - \mu )^2 + P_3(\mu_3 - \mu)^2
\end{equation}
with the probabilities $P_1$, $P_2$ and $P_3$ denoting the probabilities and $\mu_1$, $\mu_2$ and $\mu_3$ denoting the means of $C_1$, $C_2$ and $C_3$ respectively. $p_i$ is probability function for each pixel value that is based on the computed and normalized image histogram:
\begin{equation}
    {P_1{=}\sum_{i=0}^{k_1}p_i},\  {P_2{=}\sum_{i=k_1+1}^{k2}p_i},\  {P_3{=}\sum_{i=k_2+1}^{L}p_i}. 
\end{equation}
The means means for the three classes are given by:
\begin{equation}
     {\mu_1{=}\frac{1}{P_1}\sum_{i=0}^{k1}ip_i},\ {\mu_2{=}\frac{1}{P_2}\sum_{i=k_1+1}^{k2}ip_i},\ {\mu_3{=}\frac{1}{P_3}\sum_{i=k_2+1}^{L}ip_i}.
\end{equation}
The optimal thresholds can then be selected by $k^*_1$ and $k^*_2$ by maximising:
\begin{equation}
    \sigma^2_B(k^*_1,k^*_2) = \max_{1 \leq k_1 \leq k_2 \leq L} \sigma^2_B(k_1,k_2)
\end{equation}
Using the lowest threshold obtained by the multi-Otsu thresholding with three classes, $k_1$, we can then obtain a binary segmentation map of each patch by setting all values above this threshold to 0 and all below (darker-crack-pixels) to 1. 
To filter out noise of darker regions that do not belong to cracks we keep only the regions where the segmentation maps of patches are in agreement: only when all overlapping patches agree that a single pixel belongs to a crack, it is kept. This leads to a binary map of the segmented crack image in which, due to the thresholding, the cracks are surrounded by large black regions. \autoref{fig:thresholding_options} shows examples of performing different thresholding options including ours, two-class Otsu, multi-Otsu and local options such as Niblack \cite{niblack1985IntroductionDigital} and Sauvola \cite{sauvola2000AdaptiveDocument} thresholding.
This thresholded image and the coarse localization map are then merged using simple multiplication. We use the characteristic that our thresholding method leads to little noise in regions surrounding a crack and use the localization map to filter out the incorrect, thresholded segmentation results in other parts of the image. This leads to a confidence map, as the binarized, thresholded map has now been influenced by the localization map which contains confidence scores.
We then pass the remaining output through another bilateral filter to further remove noise and perform a morphological closing operation to fill any holes that may have appeared using the thresholding on the cracks. \autoref{fig:thresholding} shows the interim output maps of those steps on a sample image. 

\subsection{Pseudo-Labeling and End-To-End Segmentation Training}
With the aforementioned steps, it is possible to create pseudo-labels of images, that previously only had classification labels. This is achieved by using the dataset that has been used for classifier training and passing this through our proposed method. This will yield labels that are not as accurate as ``proper" segmentation labels but will suffice for training of an end-to-end segmentation algorithm. This is due to the batch-based training and generalization capabilities of CNN algorithms.
The right image in \autoref{fig:thresholding} shows an example of a label generated using our method and the original segmentation label for a training image. It is to note that in some instances, the pseudo-labels may have incorrectly missed a crack and not predicted anything. Therefore, to train the segmentation algorithm effectively, we propose that only label-image combinations are kept where at least 100 pixels were predicted as belonging to a crack. 

Additionally, it is also possible to extend this method for further semi-supervised learning; If not only a labelled subset is available, but a further un-labeled set, this may also be incorporated into the training process. Pseudo-labels can be generated using our method the same way as for the training set and the segmentation algorithms can then be trained on the whole, combined, dataset.

\section{Experiments}
This section contains details about the data, metrics and implementation details used for our approach.
\subsection{Data}
We use several datasets to show the performance of our weakly-supervised method. 

The \textit{CRACK500} dataset by Yang \textit{et al.} \cite{yang2019FeaturePyramid} is used for training and testing. It consists of 1896 training images and 1124 testing images. This data has been collected on a variety of surfaces on the Temple University using a smartphone. The original sized images were cropped to varying sizes of $648{\times}848$ and $640{\times}360$ pixels and all contain at least 1,000 pixels belonging to cracks

We use the training subset of this dataset for training our classifier and fully-supervised segmentation CNNs. To increase the total amount of data available we augment those images by rotating them to [$0^{\circ},+90^{\circ}$] and flip them along the vertical and horizontal axis. This yields an 8-fold increase of the training data. 
The testing subset of this dataset is used for evaluation and not augmentation is performed during test-time. 

2) Crackforest (\textit{CFD}) is a dataset proposed by Shi \textit{et al.}. It consists of 117 images\footnote{Image \textit{042.jpg} contains an incorrect ground truth and has been omitted.}. The images are of size $320\times480$ pixels and has been captured using a smartphone on road surfaces in Beijing. This dataset is used for evaluation only.

3) The images from AIGLE, ESAR and LCMS (\textit{AEL}) datasets \cite{chambon2011AutomaticRoad} are combined to provide a further evaluation set. This consists of a total of 58 images of varying sizes. That data was acquired from road surfaces and provides a variety in quality as different acquisition systems have been used. This dataset is used for evaluation only.

4) The work by Liu \textit{et al.} \cite{liu2019DeepCrackDeepa} has provided the DeepCrack (\textit{DCD}) dataset. DCD has an official train-test split of 300 and 237 images respectively. However, similar to \cite{qu2021CrackDetection}, we use the entirety of this dataset for evaluation. The data consists of cracks on a variety of surfaces such as walls, concrete and pavement.

All of the above mentioned datasets contain pixel level, segmentation annotations. However, since our method makes use of weakly-supervised learning by using classification annotations, we convert the training data to classification labels using a sliding window based, patch-extraction approach. We set the spatial size of those patches to 128 pixels, stride of patch extraction to 64 pixels and count a patch belonging to the ``crack" class if at least one pixel within the segmentation mask of that patch belongs to a crack. This is applied to augmented Crack500 training set and yields a total of around 261,000 images belonging to the ``crack" class and  436,00 images without cracks. We also apply this patch extraction approach to the Crack500 test set as we aim to observe the classification performance of the classifier which aids in selection of a suitable architecture. 

\subsection{Metrics}
Similar to \cite{konig2021OptimizedDeep, zou2018DeepCrackLearning, yang2019FeaturePyramid, qu2021CrackDetection} we use the F1-based metrics: Optimal Dataset Scale (ODS), Optimal Image Scale (OIS) and the average precision (AP). These metrics make use of the prediction pixels consisting of true positive $TP$, false positive $FP$ and false negative $FN$ predictions. To compute the OIS and ODS metrics, individual predictions $i$ are binarized at a confidence threshold $t \in [0,1]$. This is followed by computing the Recall $RE_t = TP_t / (TP_t+FN_t)$ and Precision $PR_t = TP_t / (TP_t+FP_t)$ at a threshold $t$. The F1 score $F1 = 2(RE*PR)/(RE+PR)$ is then incoporated for computation of the OIS and ODS metrics as folows:
\begin{equation}
    OIS =  \frac{1}{N_{img}}  \sum_{i}^{N_{img}}  max \Big\{ F1_t^i: \forall t \in \{0.01, ..., 0.99\} \Big\}
\end{equation}
\begin{equation}
    ODS = max \Big\{ \big\{ \frac{1}{N_{img}}  \sum_{i}^{N_{img}}   F1_t^i \big\} : \forall t \in \{0.01, ..., 0.99\} \Big\}
\end{equation}
The combination of OIS and ODS is beneficial as OIS does not overly weight images with a lot of crack pixels and ODS gives a good overview over the consistency of the algorithm across the dataset. The formula for AP is as follows:
\begin{equation}
AP = \sum_t^T \frac{1}{T} (RE_t - RE_{(t{-}0.01}))PR_t
\end{equation}
where T is the number of confidence thresholds  which is set to 100. 

\subsection{Implementation: Classifier for the Pseudo Labels}
For generation of our pseudo-labels we chose ResNet based classifiers, namely ResNet50, ResNet101 and ResNet152 \cite{he2016DeepResidual}. We modify these classifiers to only produce two output classes, as to whether a patch contains a crack or not. 
Due to the large amount of image patches that have been generated using the sliding window approach, we train the classifiers for 3 epochs. The batch size is set to 16 and the initial learning rate of $1e-3$ is reduced by a factor of 10 at every epoch. We use SGD as the optimizer and set the momentum to 0.9. The weights after training finished on last epoch are used for the next steps in our method. During training we perform image augmentation, by performing color jittering and the addition of Gaussian and multiplicative noise.

For the localization and thresholding methods in the pseudo-labelling process we set a patch extraction size of ${32{\times}32}$ pixels. The stride for the localization using the classifier is set to 16 pixels and for thresholding to 8 pixels. The parameters for both bilateral filters are set to ${\sigma_s{=}120}$, ${\sigma_r{=}}120$ and ${d{=}2}$. Both morphological operations, the erosion of the localization map and closing of the segmentation output use a kernel size of $3{\times}3$ pixels and are performed for 4 and 1 iterations respectively.

Our experiments are conducted on a deep learning workstation running Ubuntu 16.04 LTS, containing an Intel i9-7960X CPU and a Nvidia Titan XP GPU with 12GB RAM. The framework in which the experiments have been implemented is Tensorflow 2.3.

\subsection{Comparison: Pseudo-Label Generation with CRF and Class Activation Maps}
To show that our method achieves state-of-the art results in the weakly-supervised task when only using classification labels, we compare our approach with that of Dong \textit{et al.} \cite{dong2020PatchBasedWeakly}. To mirror their implementation approach, we use the trained ResNet50, ResNet101 and ResNet152 classifiers and a subset of 100 random training images and their segmentation labels to grid-search the best CRF parameters which achieve the highest mean intersection over union score for each classifier. More details about the grid-search parameters can be found in the original work \cite{dong2020PatchBasedWeakly}. We then apply this method using the best CRF parameters determined by the search on all training images to generate a set of pseudo-labels. 

\subsection{Implementation: End-To-End Segmentation CNNs}
\label{sec:e2e_seg}
To show that our method creates sufficient pseudo labels for the task of crack segmentation, we use several recent state-of-the-art crack segmentation algorithms trained on the original labels, as well as on our pseudo labels:

1) \textit{U-Net } is a encoder-decoder method initially introduced for medical image data \cite{ronneberger2015UNetConvolutional}. Here, the encoder and the decoder are connected through skip-connections which provide better feature retention. It has been previously used for crack segmentation \cite{jenkins2018DeepConvolutionala, konig2019ConvolutionalNeurala}.

2) \textit{DeepCrack} by Liu \textit{et al.}\cite{liu2019DeepCrackDeepa} is a fully convolutional method that does not have an explicit decoder part. Rather, it upscales feature maps from the encoder and uses guided filtering to create the segmentation output.

3) \textit{DeepCrack} by Zou \textit{et al.}\cite{zou2018DeepCrackLearning} is a encoder-decoder based model which fuses multi scale information from the encoder and decoder and merges them to create the segmentation output. This multi-scale fusion improves crack segmentation.

4) \textit{OED} (Optimized-Deep-Encoder-Decoder) in \cite{konig2021OptimizedDeep} uses a pretrained EfficientNet as an encoder and included a custom decoder section that has been designed for crack segmentation. This method losely follows the U-Net shape.

These segmentation methods are trained on the original pixel-annotation labels in a fully-supervised (FSV) manner as well as using our proposed pseudo-labeling, weakly-supervised (WSV) manner. Each of those algorithms is trained for a total of 30 epochs, using a batch size of 4. During training we perform the same data augmentation of color jittering and addition of noise as in the classifier training. The initial learning rate is set to $1e{-}3$ and reduced by a factor of 10 after every 10 epochs. SGD is used as the optimizer and a momentum is set to 0.9. The binary cross-entropy loss is used to compare the predicted output with the ground truth during training. Note that when using our weakly-supervised method, the ground-truth is set to be the pseudo labels. 
The predictions of those methods are performed on the full sized testing images. 

\begin{table}[!tb]
    \centering
    \caption{Classification results of the CNN classifier on the patches of the CRACK500 test using the F1 score, Precision and Recall}
    \renewcommand{\arraystretch}{0.8}
    \begin{tabular}{llll}\toprule
        Classifier & F1 & PR & RE  \\\midrule
        ResNet50 & 0.9045 & 0.8824 & \underline{0.9278} \\
        ResNet101 & \underline{0.9054} & 0.\underline{8864} & 0.9252 \\
        ResNet152 & 0.8891 & 0.8836 & 0.8948 \\
    \bottomrule
    \end{tabular}
    \label{tab:class_results}
\end{table}

\begin{table}[!tb]
    \centering
    \caption{Pseudo-label quality and creation speed in comparison with the actual segmentation labels of the training set. }
    \renewcommand{\arraystretch}{0.8}
    \begin{tabular}{lllll}\toprule
        Pseudo-Label-Type & ODS & OIS & AP & images/sec \\\midrule
        CRF-Labels \cite{dong2020PatchBasedWeakly} (ResNet50) &  0.3796 &  0.3796 &  0.1851 & \underline{0.8739}\\
        CRF-Labels \cite{dong2020PatchBasedWeakly} (ResNet101) &  0.2874 &  0.2874 &  0.1278 & 0.5029\\
        CRF-Labels \cite{dong2020PatchBasedWeakly} (ResNet152) &  0.2698 &  0.2698 &  0.1295 & 0.8266\\
        \midrule
        Our-Labels (ResNet50)  &  \underline{0.5359} & \underline{0.5839} & \underline{0.4371} & 0.1868 \\
        Our-Labels (ResNet101) &  0.5040 &  0.5706 & 0.3749 & 0.1841\\
        Our-Labels (ResNet152) &  0.5210 &  0.5690 & 0.4095 & 0.1837 \\

    \bottomrule
    \end{tabular}
    \label{tab:pseudo_label_quality}
\end{table}

\section{Results and Discussion}
This section contains the results of using our weakly-supervised method, in comparison to fully-supervised training as well as results from ablation studies. 

\begin{figure}[ht]
\centering
\captionsetup[subfigure]{labelformat=empty}
\subfloat[]{\includegraphics[width=0.24\linewidth]{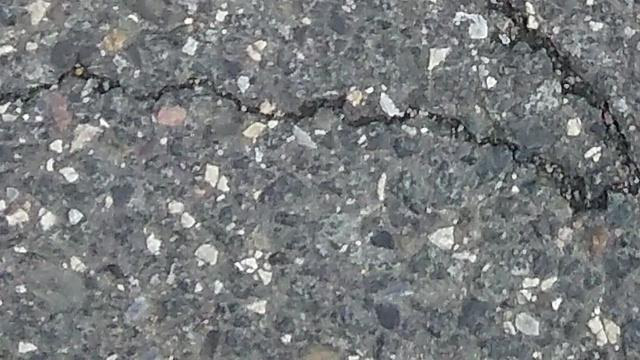}}\vspace{-0.16cm}
\subfloat[]{\includegraphics[width=0.24\linewidth]{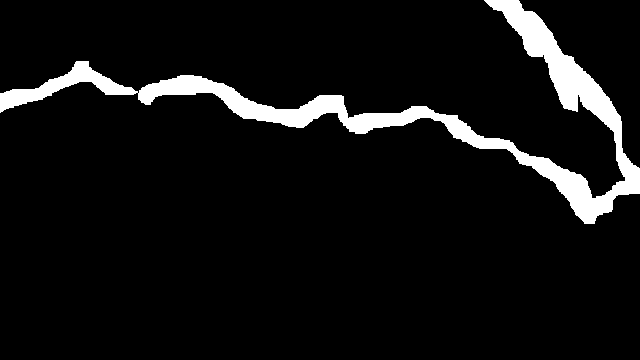}}\vspace{-0.16cm}
\subfloat[]{\includegraphics[width=0.24\linewidth]{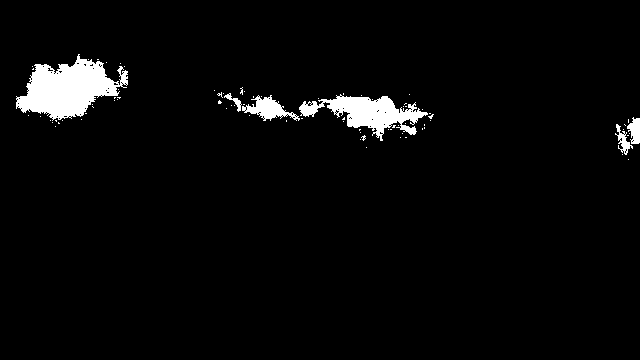}}\vspace{-0.16cm}
\subfloat[]{\includegraphics[width=0.24\linewidth]{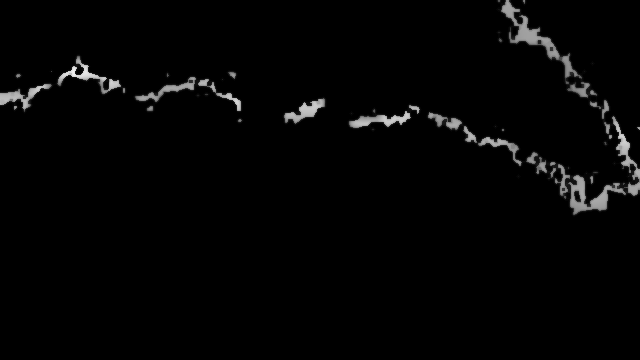}}\vspace{-0.16cm} \\
\subfloat[]{\includegraphics[width=0.24\linewidth]{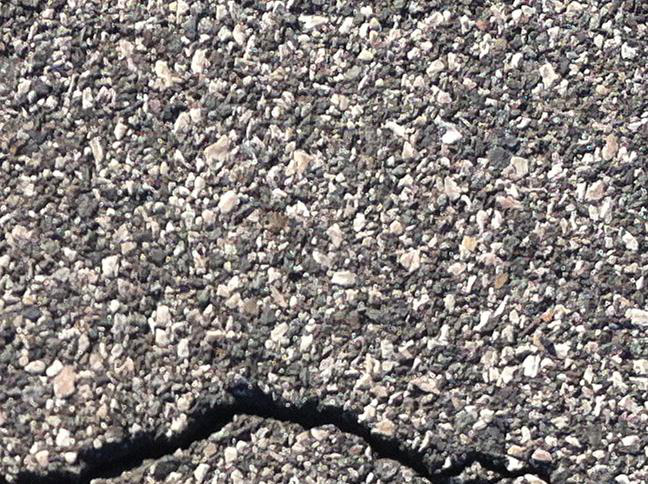}}\vspace{-0.16cm}
\subfloat[]{\includegraphics[width=0.24\linewidth]{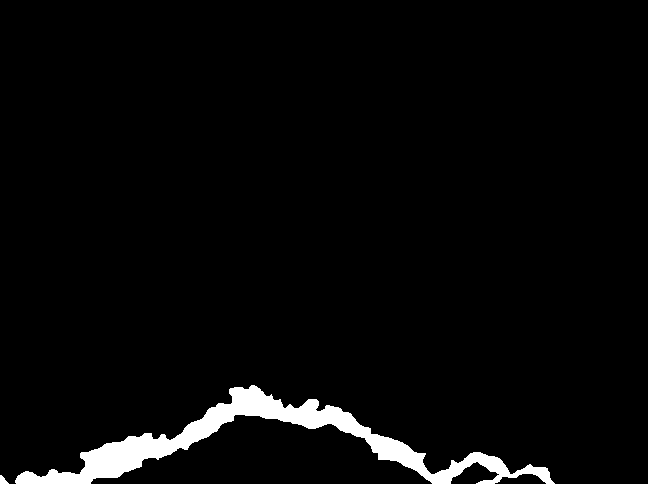}}\vspace{-0.16cm}
\subfloat[]{\includegraphics[width=0.24\linewidth]{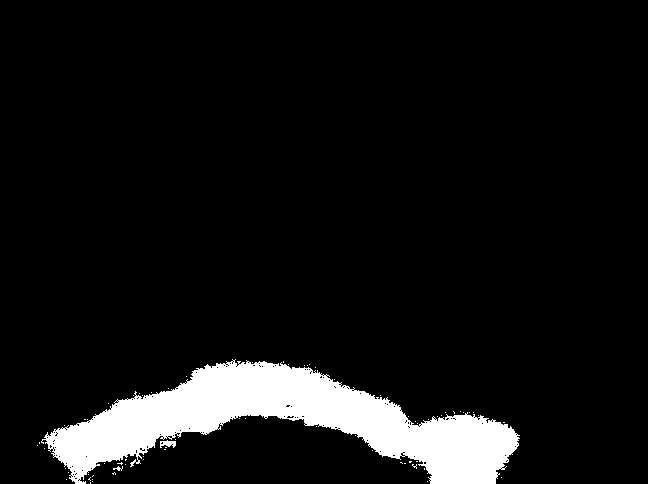}}\vspace{-0.16cm}
\subfloat[]{\includegraphics[width=0.24\linewidth]{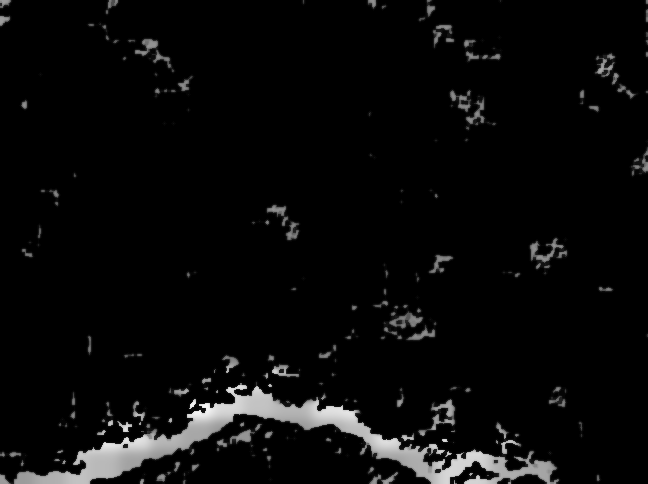}}\vspace{-0.16cm} \\
\subfloat[]{\includegraphics[width=0.24\linewidth]{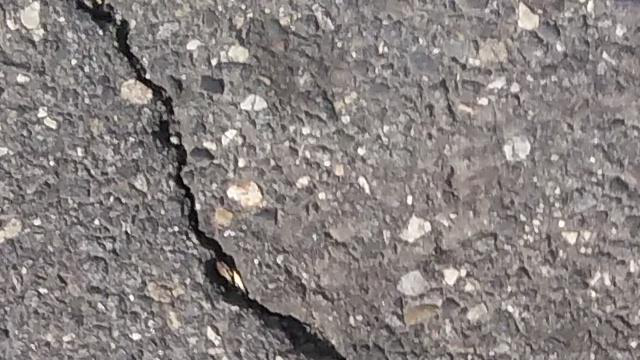}}\vspace{-0.16cm}
\subfloat[]{\includegraphics[width=0.24\linewidth]{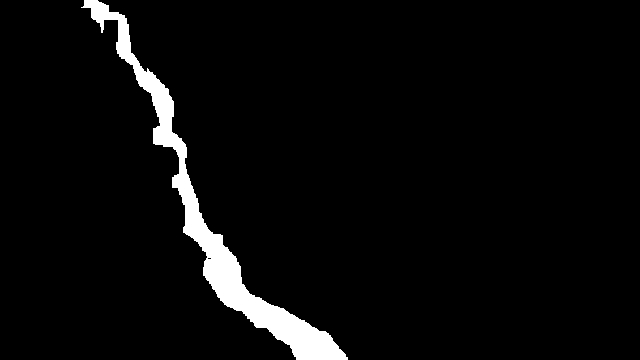}}\vspace{-0.16cm}
\subfloat[]{\includegraphics[width=0.24\linewidth]{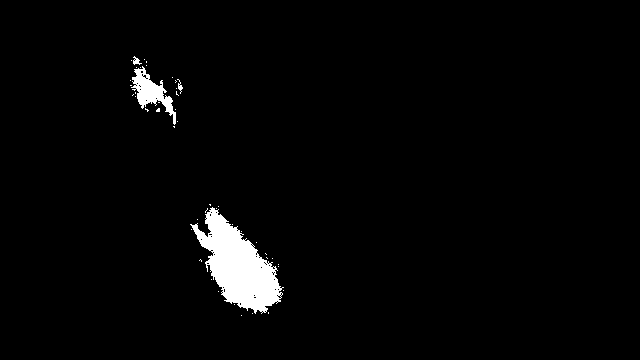}}\vspace{-0.16cm}
\subfloat[]{\includegraphics[width=0.24\linewidth]{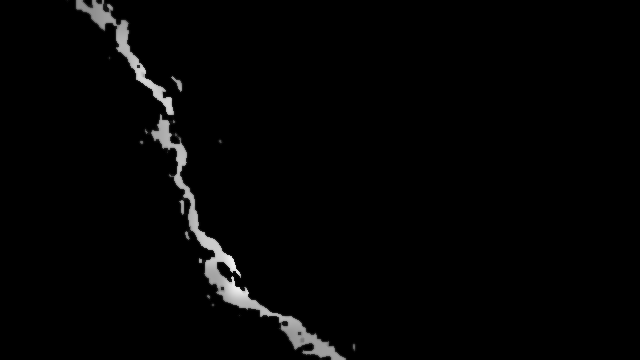}}\vspace{-0.16cm} \\
\subfloat[Image]{\includegraphics[width=0.24\linewidth]{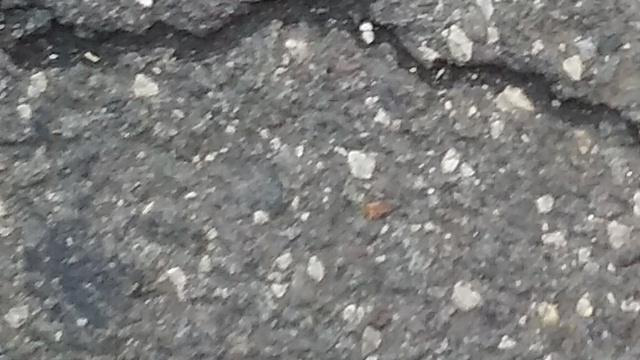}}\vspace{-0.1cm}
\subfloat[GT]{\includegraphics[width=0.24\linewidth]{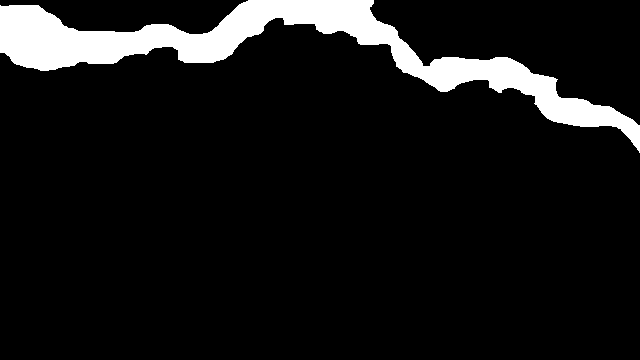}}\vspace{-0.1cm}
\subfloat[CRF Label]{\includegraphics[width=0.24\linewidth]{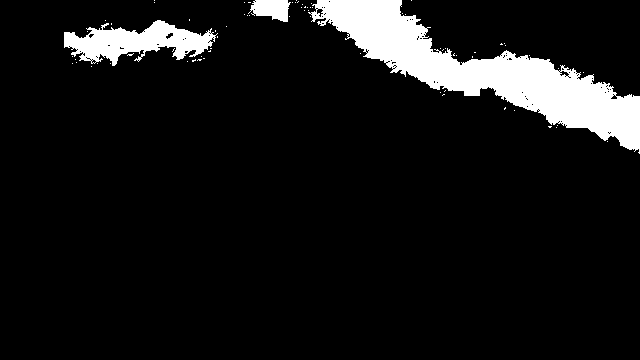}}\vspace{-0.1cm}
\subfloat[Ours Label]{\includegraphics[width=0.24\linewidth]{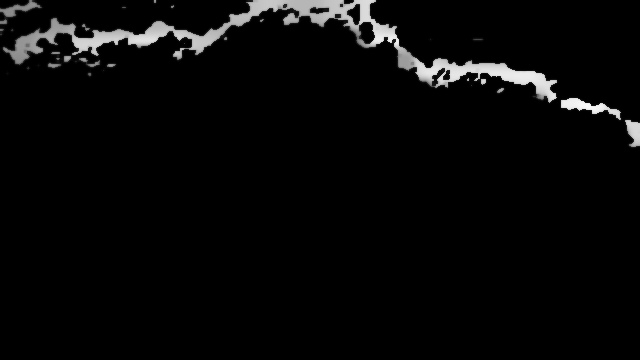}}\vspace{-0.1cm} \\
\vspace{0.3cm}
\caption{Comparison of the label quality with the ground truth of our proposed method and the CRF based method from \cite{dong2020PatchBasedWeakly}.}
\label{fig:pseudo_label_comparison}
\end{figure}
\subsection{Classifier Results and Pseudo-Label Quality}
\autoref{tab:class_results} shows the results of the classification performance of the three ResNet architectures that have been trained.
The results showing quality of the pseudo-labels in comparison with the  original segmentation labels of the CRACK500 training set is shown in \autoref{tab:pseudo_label_quality}. Here, the images/sec reports the average speed when creating pseudo-labels on the CRACK500 dataset using our aforementioned hard and software configuration. 
As it can be seen ResNet101 performs the highest in the classification task on the CRACK500 test set, performing slightly better than ResNet50. 
However, it can be seen that the label quality of our method when using the ResNet50 approach is higher than using the other classifier backbones. Due to this, in conjunction with the only slightly lower performance in the classification task and a slightly faster approach for generating the pseudo labels, we chose the ResNet50 for generating pseudo labels in the following sections. Additionally, our method creates better pseudo labels than the approach that uses CRF on the gradient activation maps by Dong \textit{et al.} independent from the classifier backbones used. 
Sample pseudo labels for the CRF method and our method using a ResNet50 classifier are shown in \autoref{fig:pseudo_label_comparison}.

\begin{figure*}[!ht]
    \scriptsize
    \centering
    \setlength{\tabcolsep}{1pt}
    \renewcommand{\arraystretch}{0.8}
    \newcommand{\centered}[1]{\begin{tabular}{l} #1 \end{tabular}}
 \newcolumntype{B}{>{\centeringrraybackslash} m{0.18\linewidth}}
\begin{tabular}{lllllllll}

\vspace{0.5em}
\rotatebox[origin=c]{90}{CRACK500} &
\raisebox{-.5\height}{\includegraphics[width=8em]{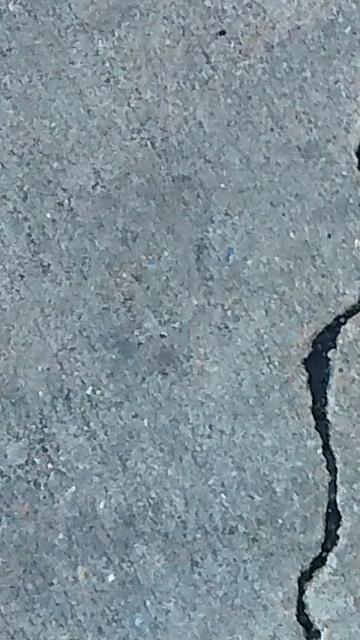} } &
\raisebox{-.5\height}{\includegraphics[width=8em]{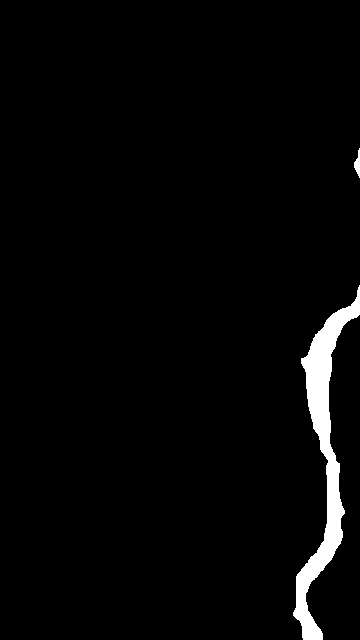} }&
\raisebox{-.5\height}{\includegraphics[width=8em]{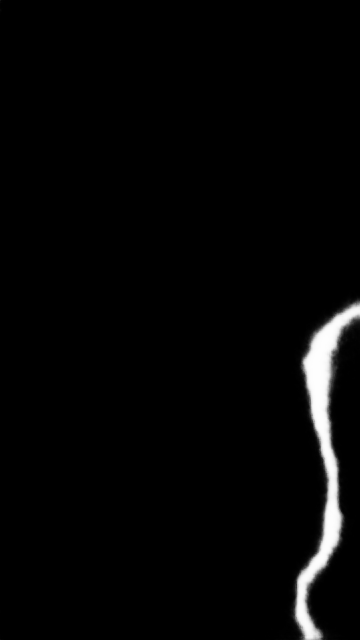} }&
\raisebox{-.5\height}{\includegraphics[width=8em]{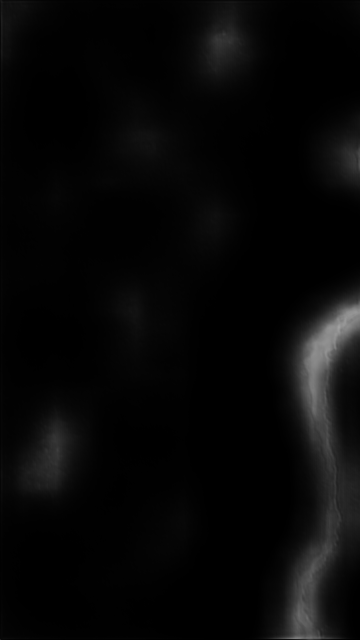} }&
\raisebox{-.5\height}{\includegraphics[width=8em]{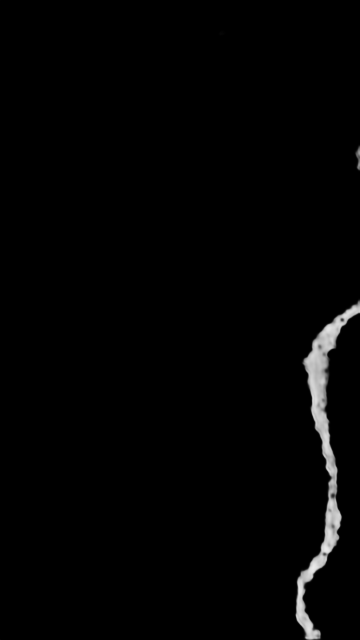} }&
\raisebox{-.5\height}{\includegraphics[width=8em]{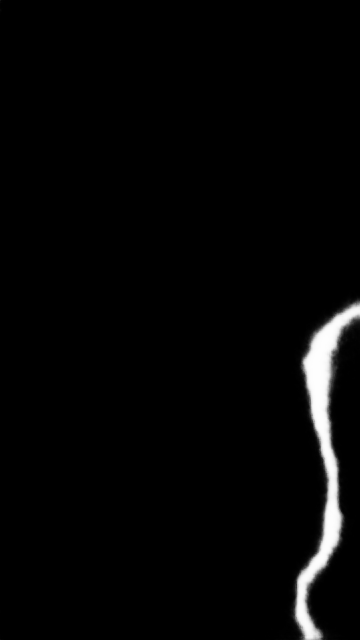} }&
\raisebox{-.5\height}{\includegraphics[width=8em]{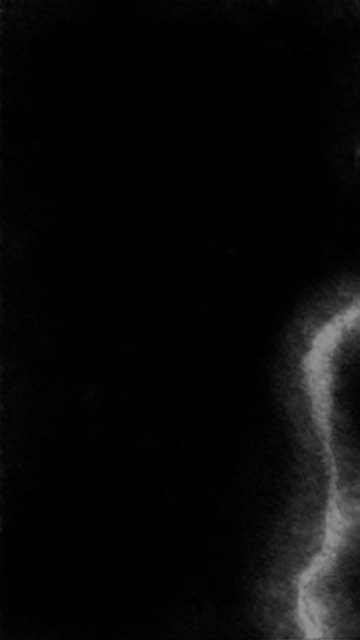} }&
\raisebox{-.5\height}{\includegraphics[width=8em]{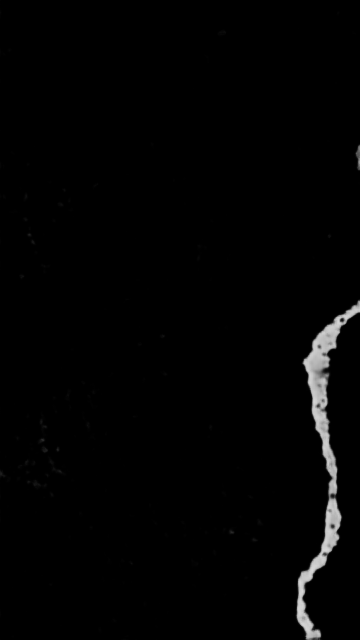} }\\ 
\vspace{1em}
\rotatebox[origin=c]{90}{CRACK500} &
\raisebox{-.5\height}{\includegraphics[width=8em]{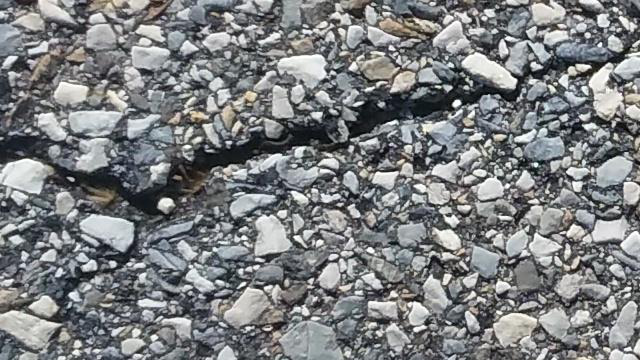} } &
\raisebox{-.5\height}{\includegraphics[width=8em]{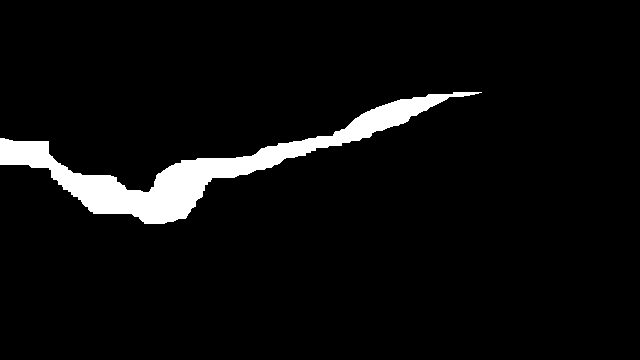} }&
\raisebox{-.5\height}{\includegraphics[width=8em]{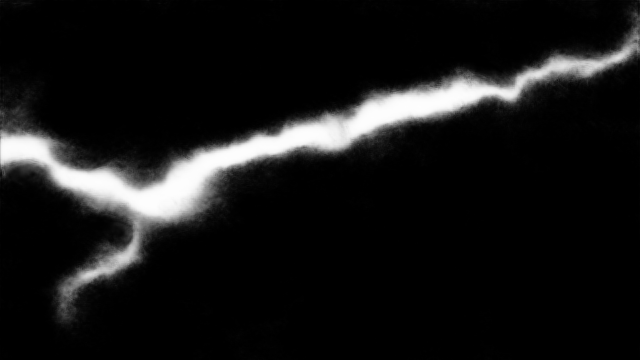} }&
\raisebox{-.5\height}{\includegraphics[width=8em]{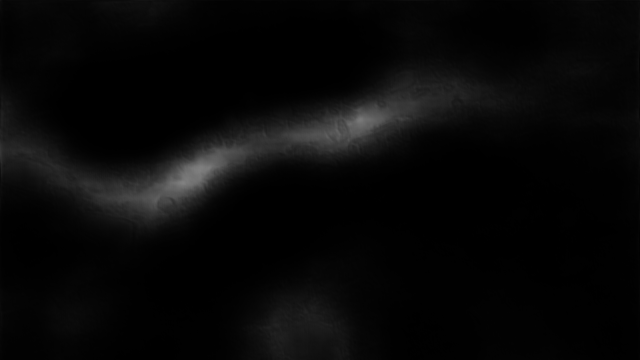} }&
\raisebox{-.5\height}{\includegraphics[width=8em]{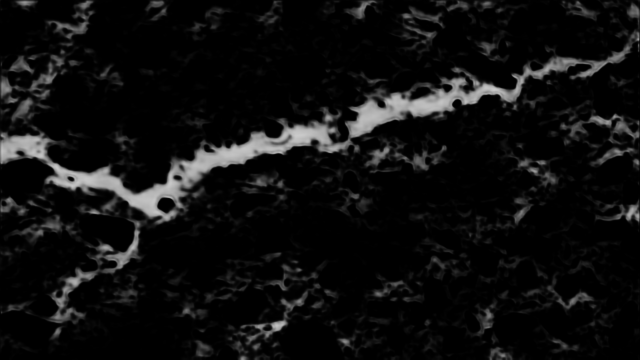} }&
\raisebox{-.5\height}{\includegraphics[width=8em]{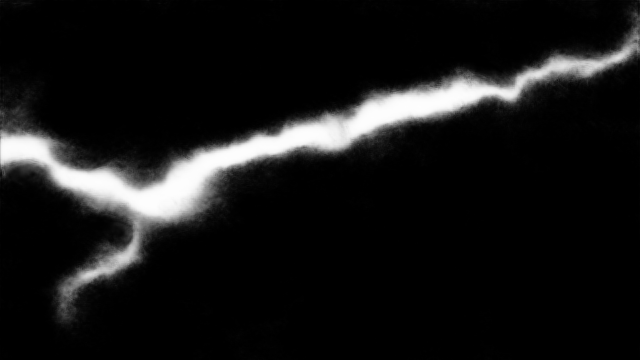} }&
\raisebox{-.5\height}{\includegraphics[width=8em]{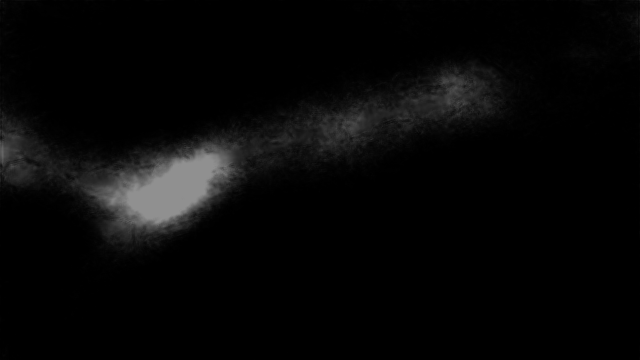} }&
\raisebox{-.5\height}{\includegraphics[width=8em]{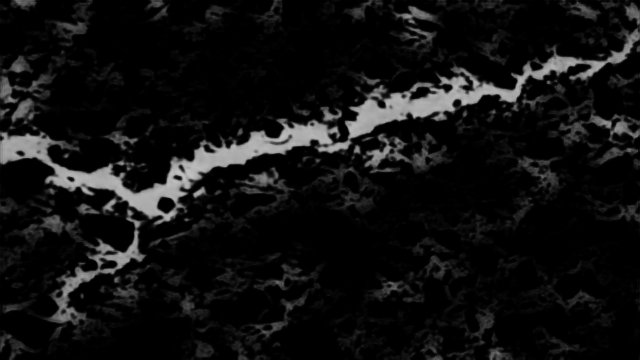} }\\ 
\vspace{1em}
\rotatebox[origin=c]{90}{AEL} &
\raisebox{-.5\height}{\includegraphics[width=8em]{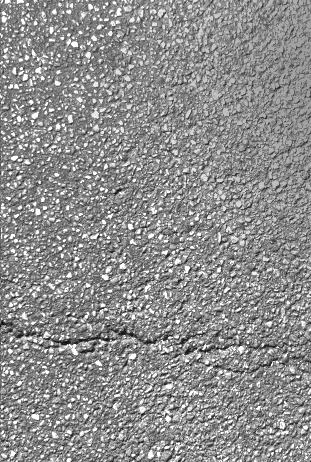} } &
\raisebox{-.5\height}{\includegraphics[width=8em]{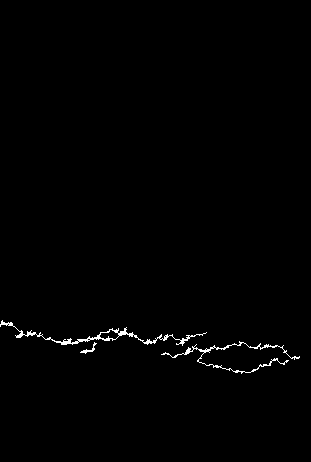} }&
\raisebox{-.5\height}{\includegraphics[width=8em]{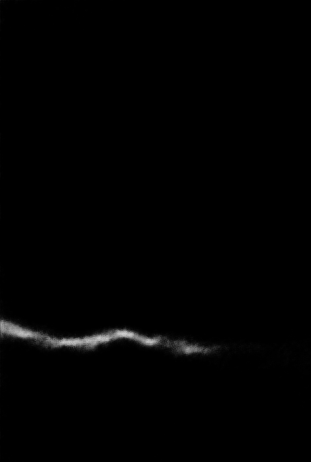} }&
\raisebox{-.5\height}{\includegraphics[width=8em]{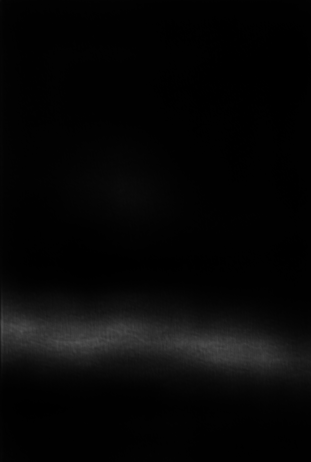} }&
\raisebox{-.5\height}{\includegraphics[width=8em]{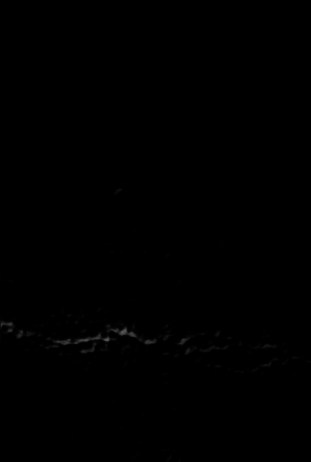} }&
\raisebox{-.5\height}{\includegraphics[width=8em]{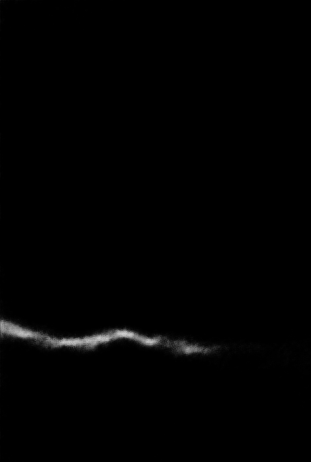} }&
\raisebox{-.5\height}{\includegraphics[width=8em]{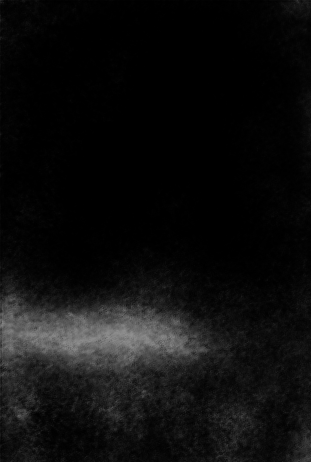} }&
\raisebox{-.5\height}{\includegraphics[width=8em]{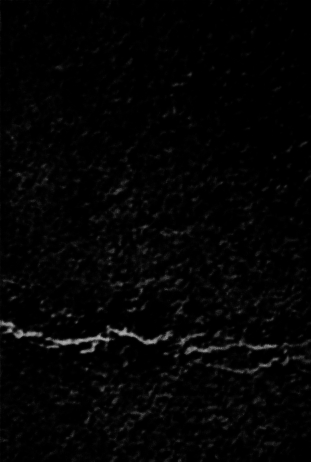} }\\ 
\vspace{1em}
\rotatebox[origin=c]{90}{AEL} &
\raisebox{-.5\height}{\includegraphics[width=8em]{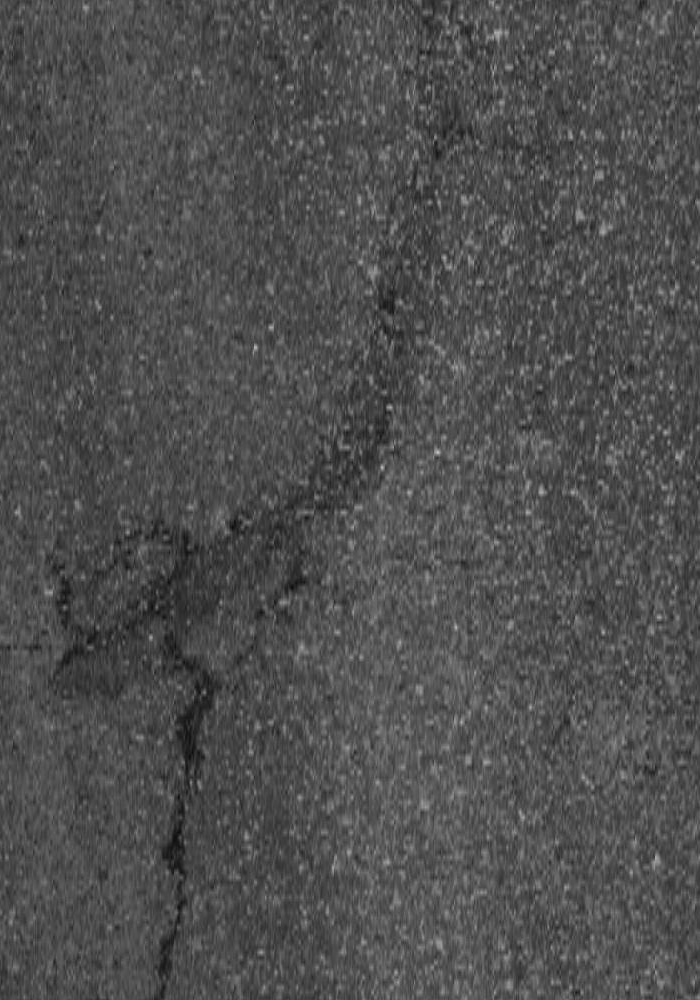} } &
\raisebox{-.5\height}{\includegraphics[width=8em]{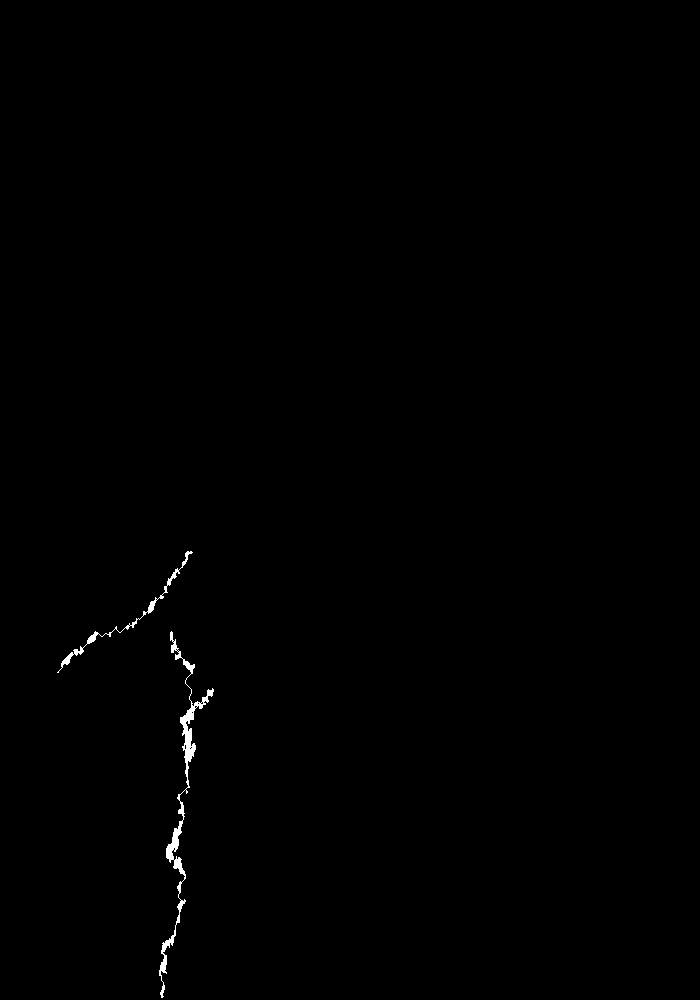} }&
\raisebox{-.5\height}{\includegraphics[width=8em]{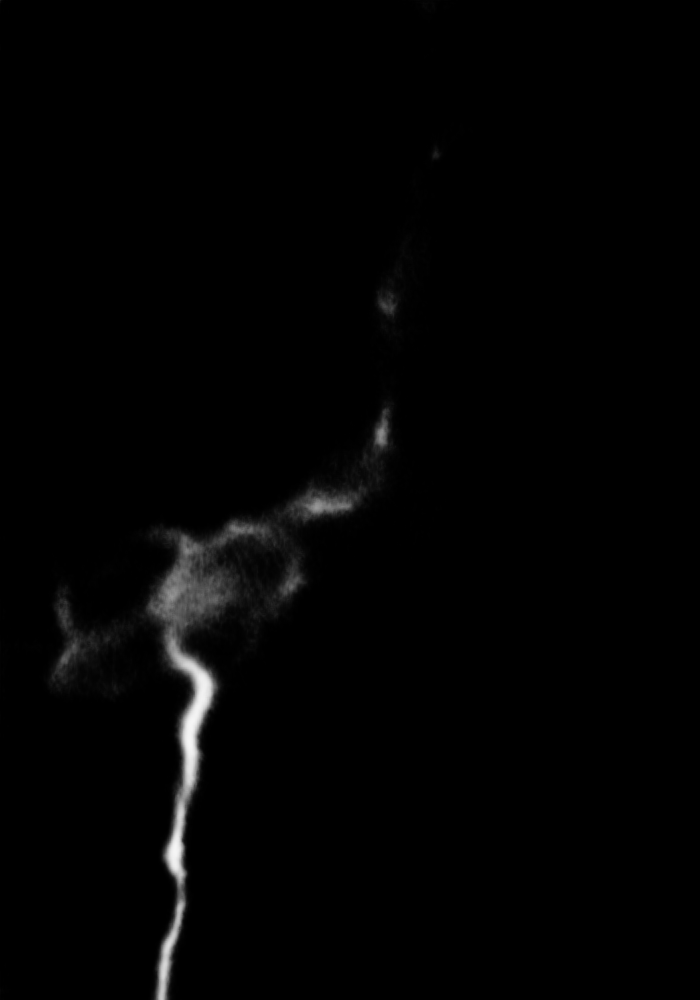} }&
\raisebox{-.5\height}{\includegraphics[width=8em]{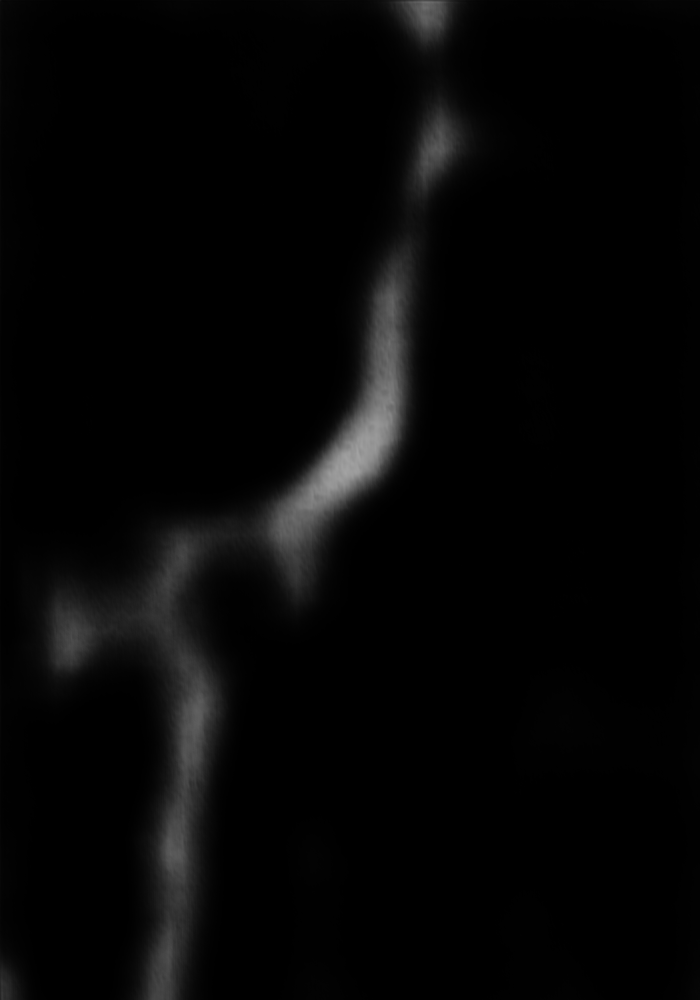} }&
\raisebox{-.5\height}{\includegraphics[width=8em]{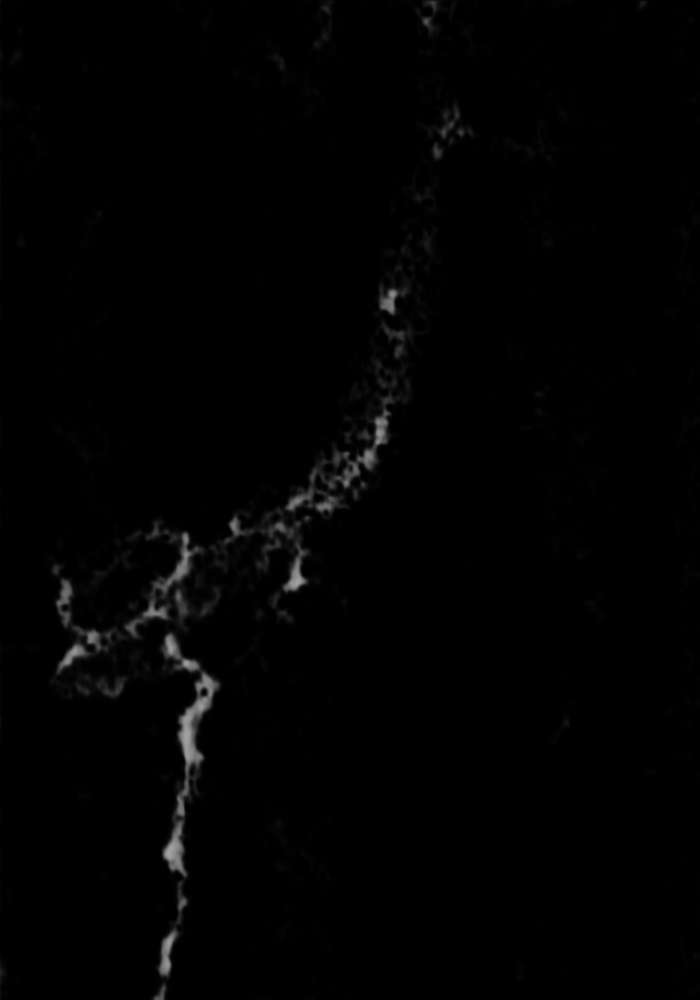} }&
\raisebox{-.5\height}{\includegraphics[width=8em]{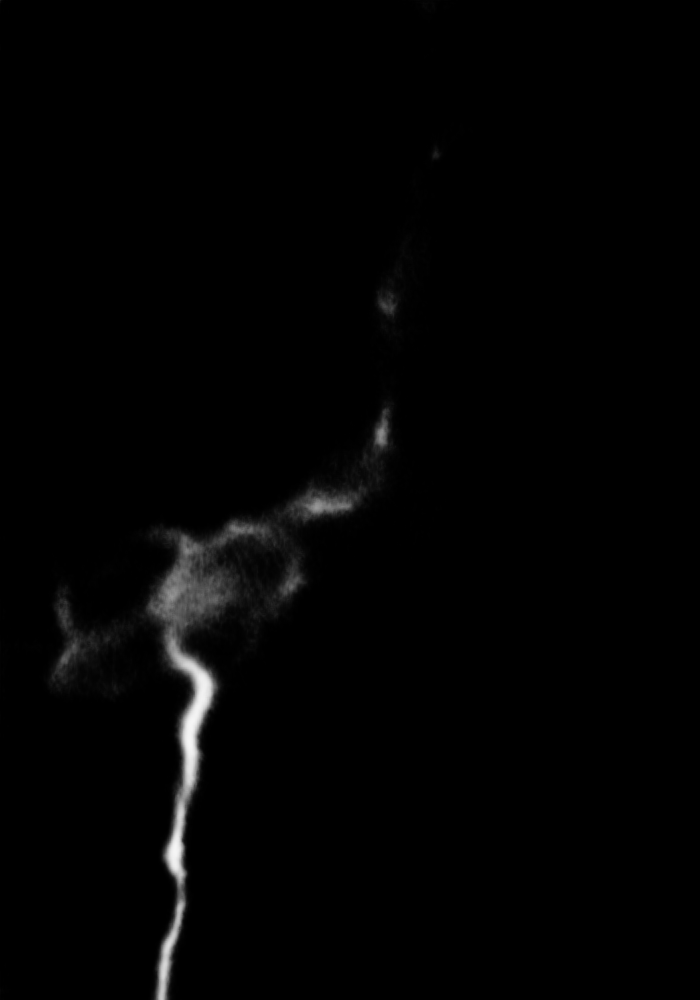} }&
\raisebox{-.5\height}{\includegraphics[width=8em]{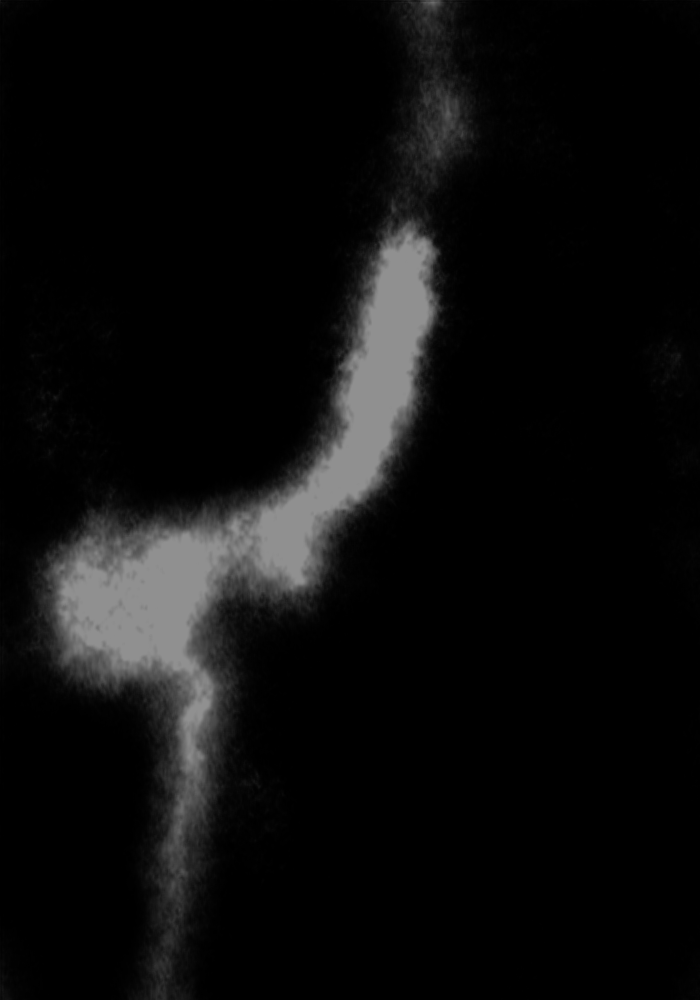} }&
\raisebox{-.5\height}{\includegraphics[width=8em]{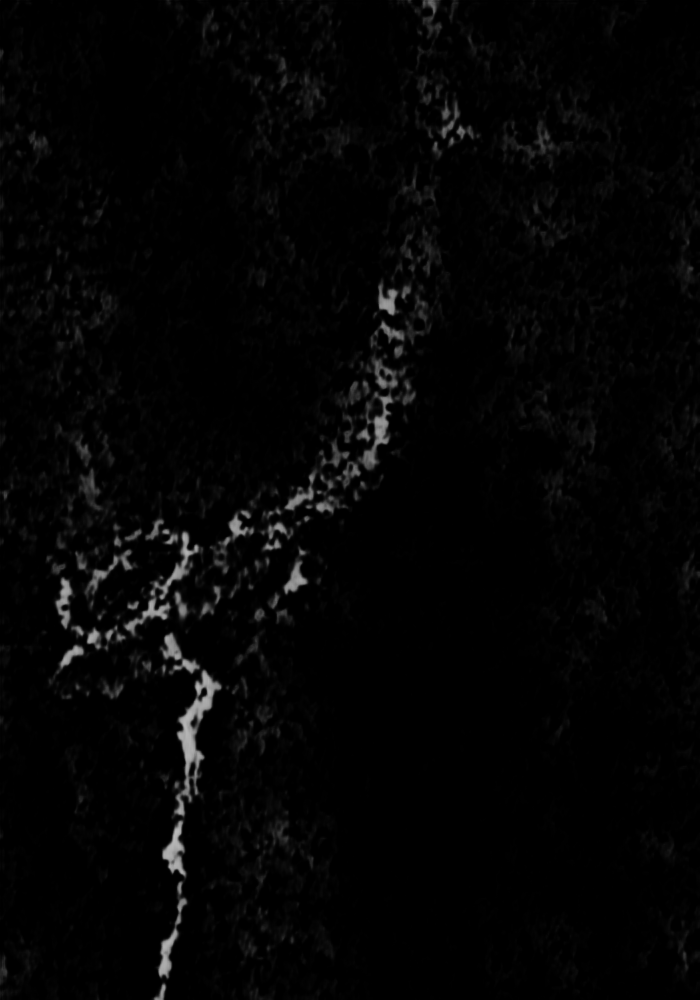} }\\ 
\vspace{1em}
\rotatebox[origin=c]{90}{CFD} &
\raisebox{-.5\height}{\includegraphics[width=8em]{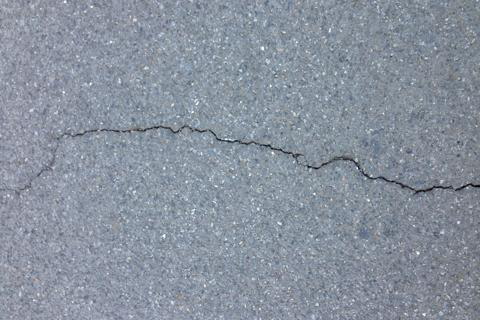} } &
\raisebox{-.5\height}{\includegraphics[width=8em]{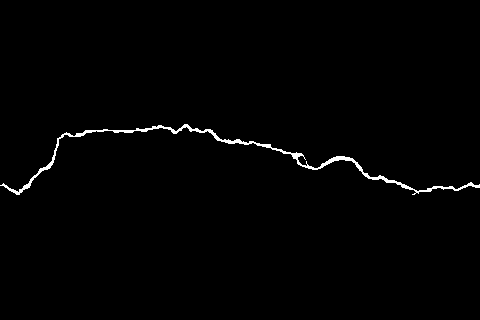} }&
\raisebox{-.5\height}{\includegraphics[width=8em]{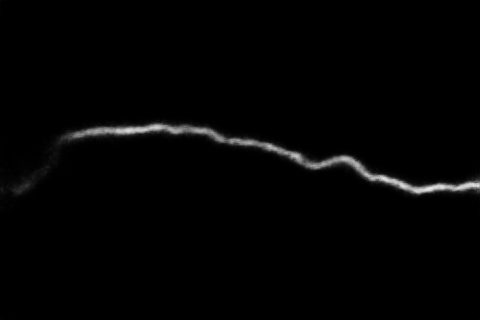} }&
\raisebox{-.5\height}{\includegraphics[width=8em]{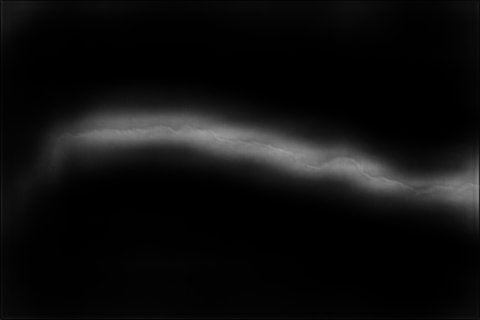} }&
\raisebox{-.5\height}{\includegraphics[width=8em]{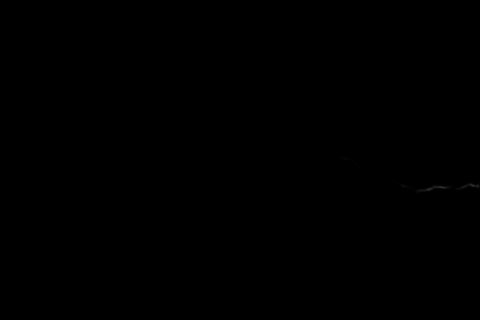} }&
\raisebox{-.5\height}{\includegraphics[width=8em]{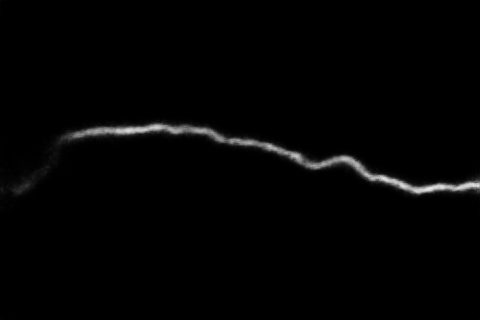} }&
\raisebox{-.5\height}{\includegraphics[width=8em]{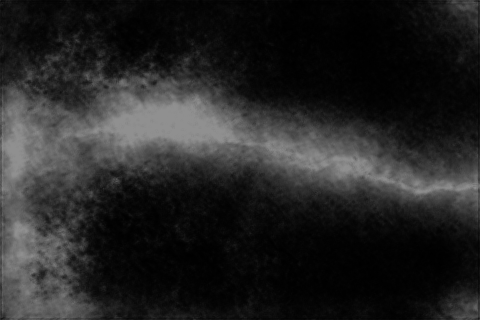} }&
\raisebox{-.5\height}{\includegraphics[width=8em]{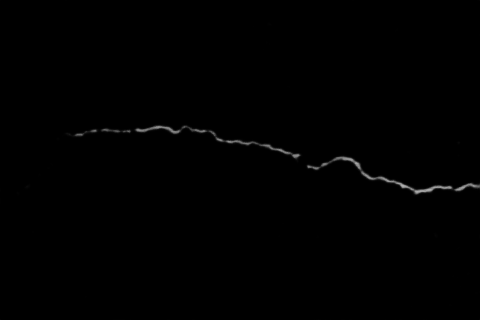} }\\ 
\vspace{1em}
\rotatebox[origin=c]{90}{CFD} &
\raisebox{-.5\height}{\includegraphics[width=8em]{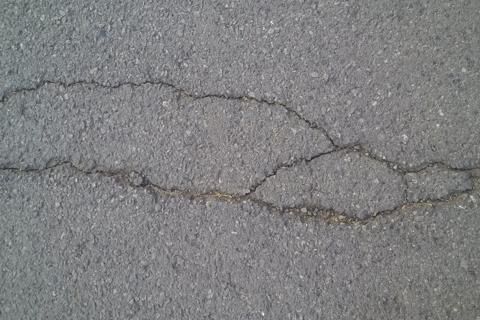} } &
\raisebox{-.5\height}{\includegraphics[width=8em]{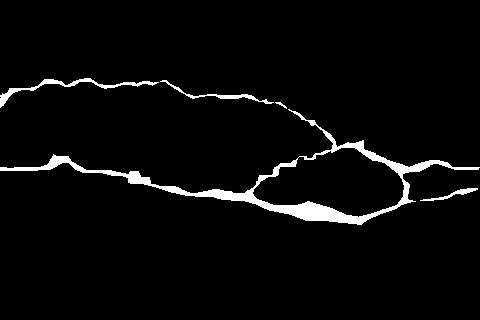} }&
\raisebox{-.5\height}{\includegraphics[width=8em]{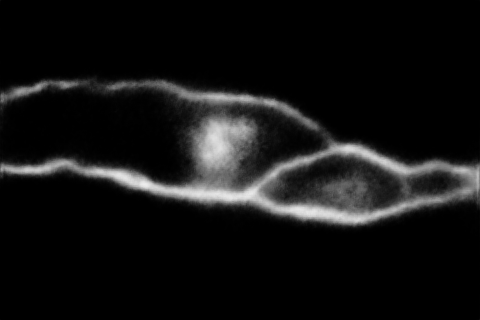} }&
\raisebox{-.5\height}{\includegraphics[width=8em]{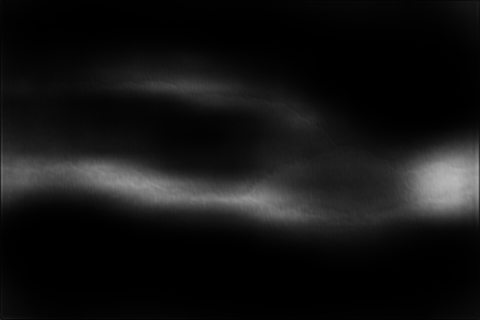} }&
\raisebox{-.5\height}{\includegraphics[width=8em]{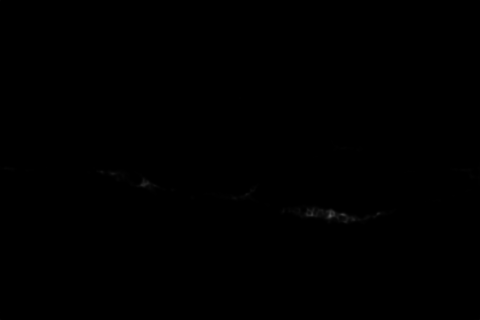} }&
\raisebox{-.5\height}{\includegraphics[width=8em]{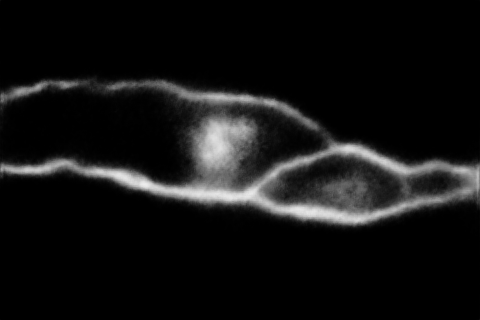} }&
\raisebox{-.5\height}{\includegraphics[width=8em]{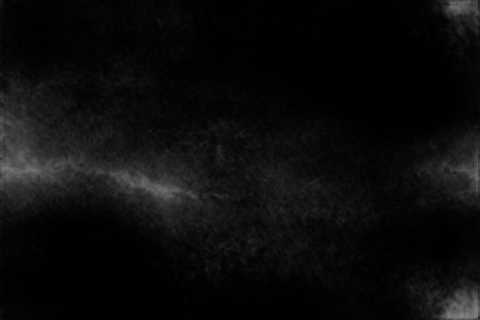} }&
\raisebox{-.5\height}{\includegraphics[width=8em]{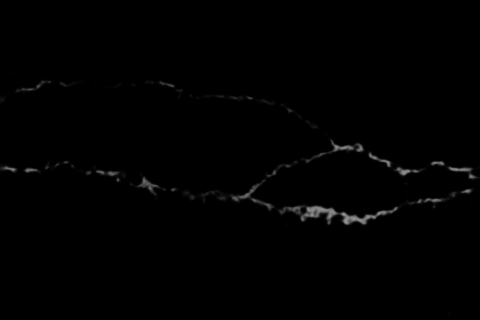} }\\ 
\vspace{1em}
\rotatebox[origin=c]{90}{DCD} &
\raisebox{-.5\height}{\includegraphics[width=8em]{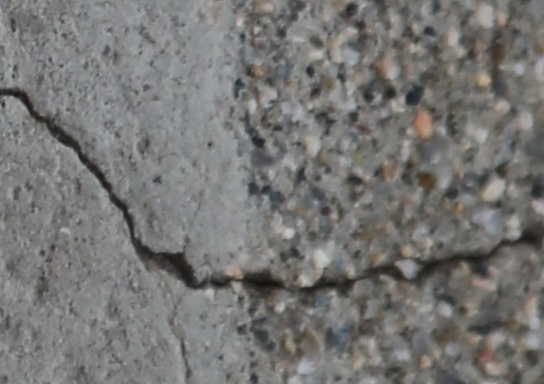} } &
\raisebox{-.5\height}{\includegraphics[width=8em]{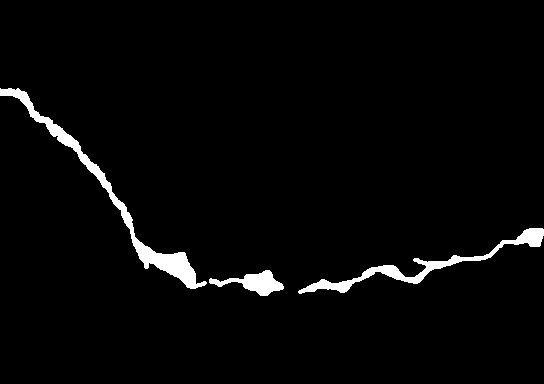} }&
\raisebox{-.5\height}{\includegraphics[width=8em]{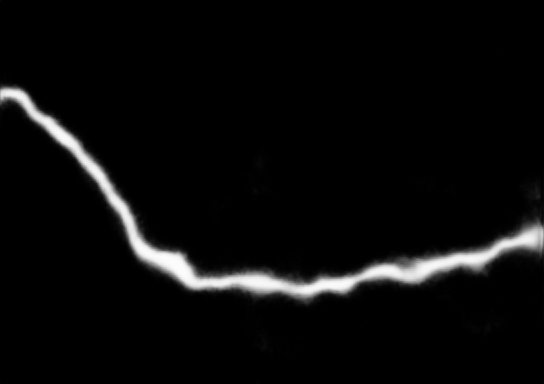} }&
\raisebox{-.5\height}{\includegraphics[width=8em]{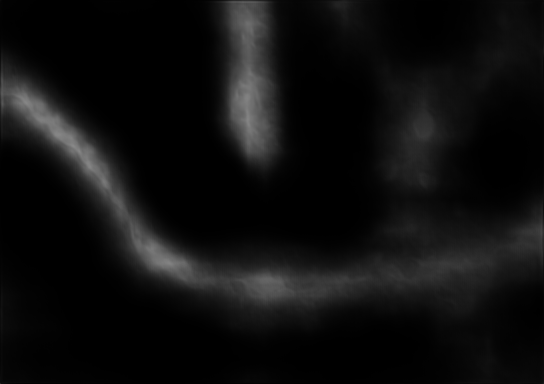} }&
\raisebox{-.5\height}{\includegraphics[width=8em]{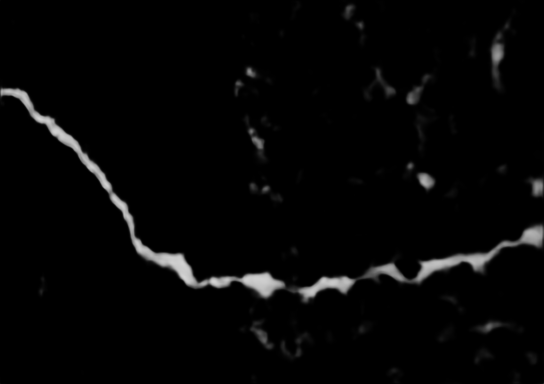} }&
\raisebox{-.5\height}{\includegraphics[width=8em]{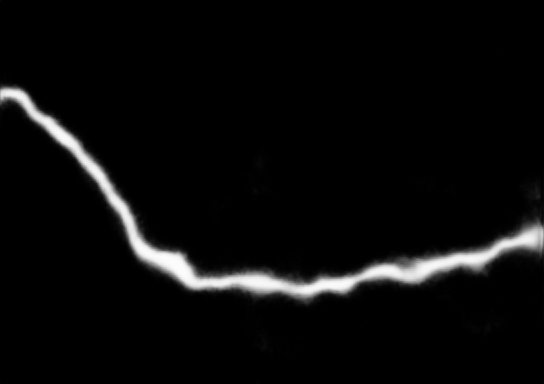} }&
\raisebox{-.5\height}{\includegraphics[width=8em]{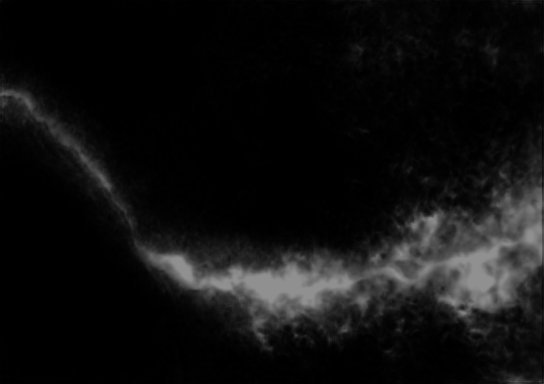} }&
\raisebox{-.5\height}{\includegraphics[width=8em]{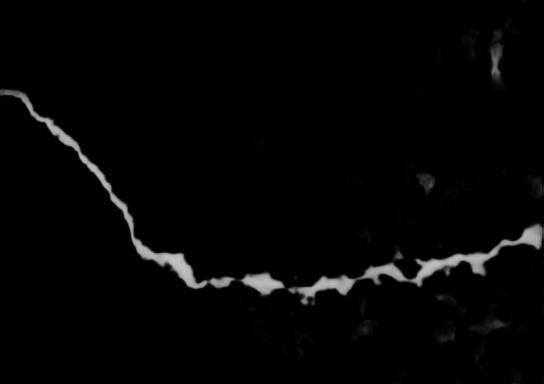} }\\ 
\vspace{1em}
\rotatebox[origin=c]{90}{DCD} &
\raisebox{-.5\height}{\includegraphics[width=8em]{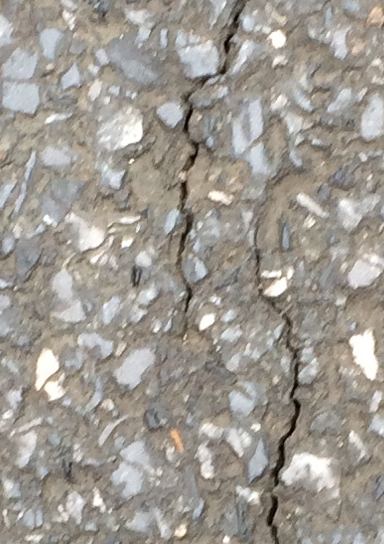} } &
\raisebox{-.5\height}{\includegraphics[width=8em]{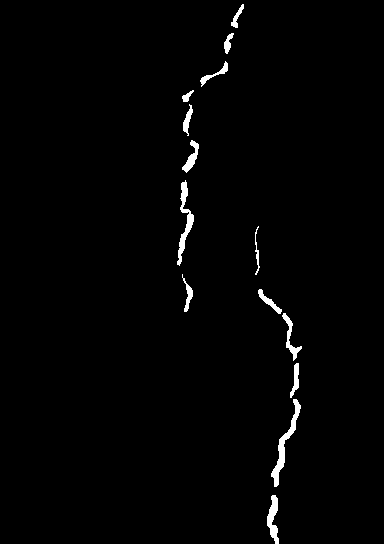} }&
\raisebox{-.5\height}{\includegraphics[width=8em]{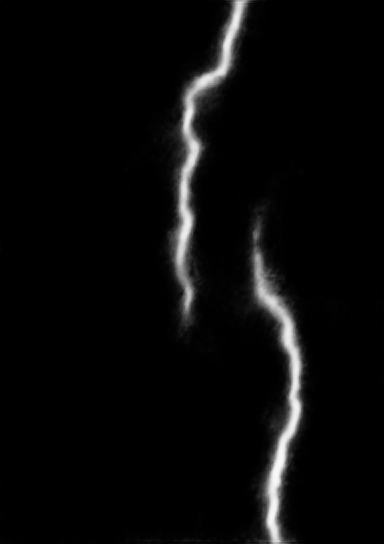} }&
\raisebox{-.5\height}{\includegraphics[width=8em]{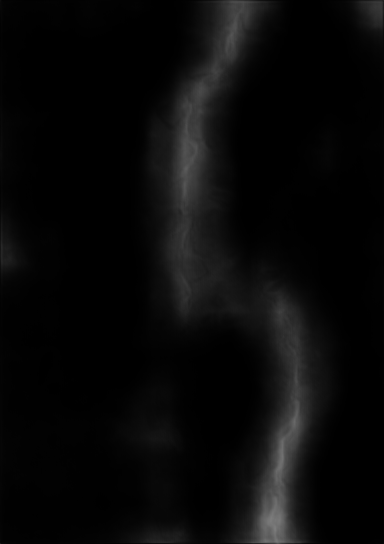} }&
\raisebox{-.5\height}{\includegraphics[width=8em]{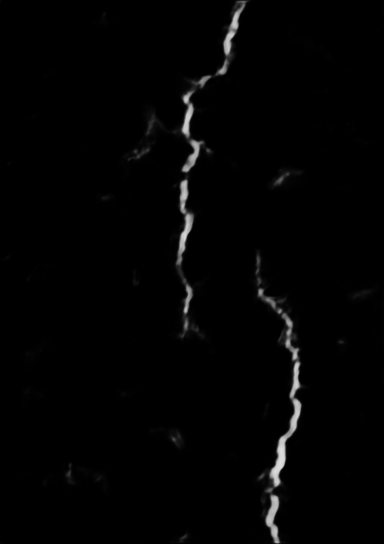} }&
\raisebox{-.5\height}{\includegraphics[width=8em]{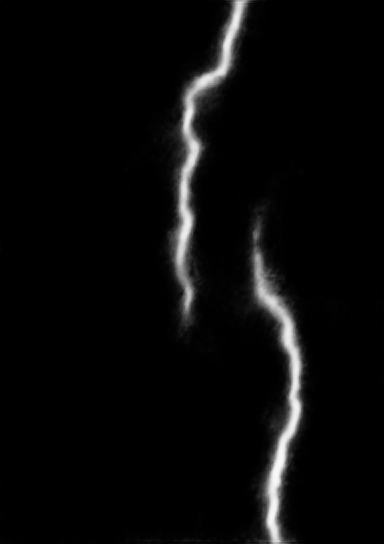} }&
\raisebox{-.5\height}{\includegraphics[width=8em]{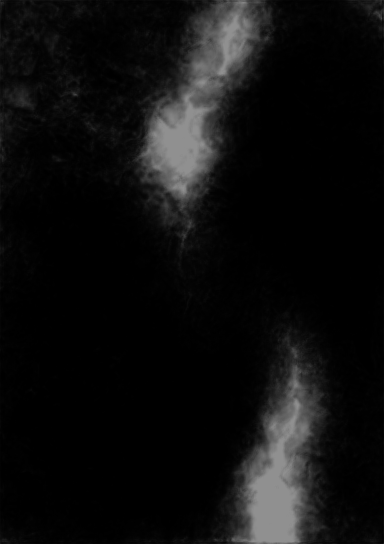} }&
\raisebox{-.5\height}{\includegraphics[width=8em]{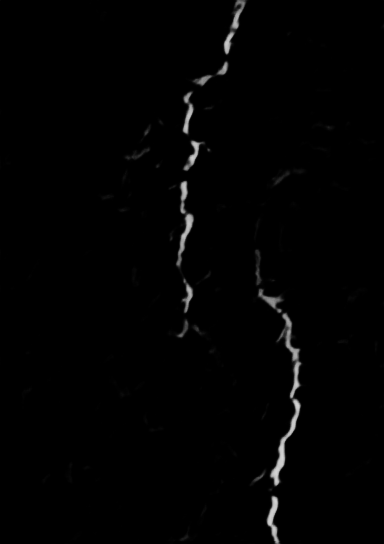} }\\

& IMG & GT & FSV: & WSV (CRF \cite{dong2020PatchBasedWeakly}): & WSV (Ours): & FSV: & WSV (CRF \cite{dong2020PatchBasedWeakly}): & WSV (Ours): \\
&  &  &  Deepcrack \cite{zou2018DeepCrackLearning} & Deepcrack \cite{zou2018DeepCrackLearning} & Deepcrack \cite{zou2018DeepCrackLearning} & OED \cite{konig2021OptimizedDeep} & OED \cite{konig2021OptimizedDeep}& OED \cite{konig2021OptimizedDeep}\\

 \end{tabular}
    \caption{Visualisation results of the outputs of training the OED \cite{konig2021OptimizedDeep} and Deepcrack \cite{zou2018DeepCrackLearning} methods trained fully supervised (FSV), weakly supervised (WSV) in comparison}
    \label{fig:results}
\end{figure*}

\subsection{Segmentation Results}
\begin{figure*}
    \centering
    \includegraphics[width=\linewidth]{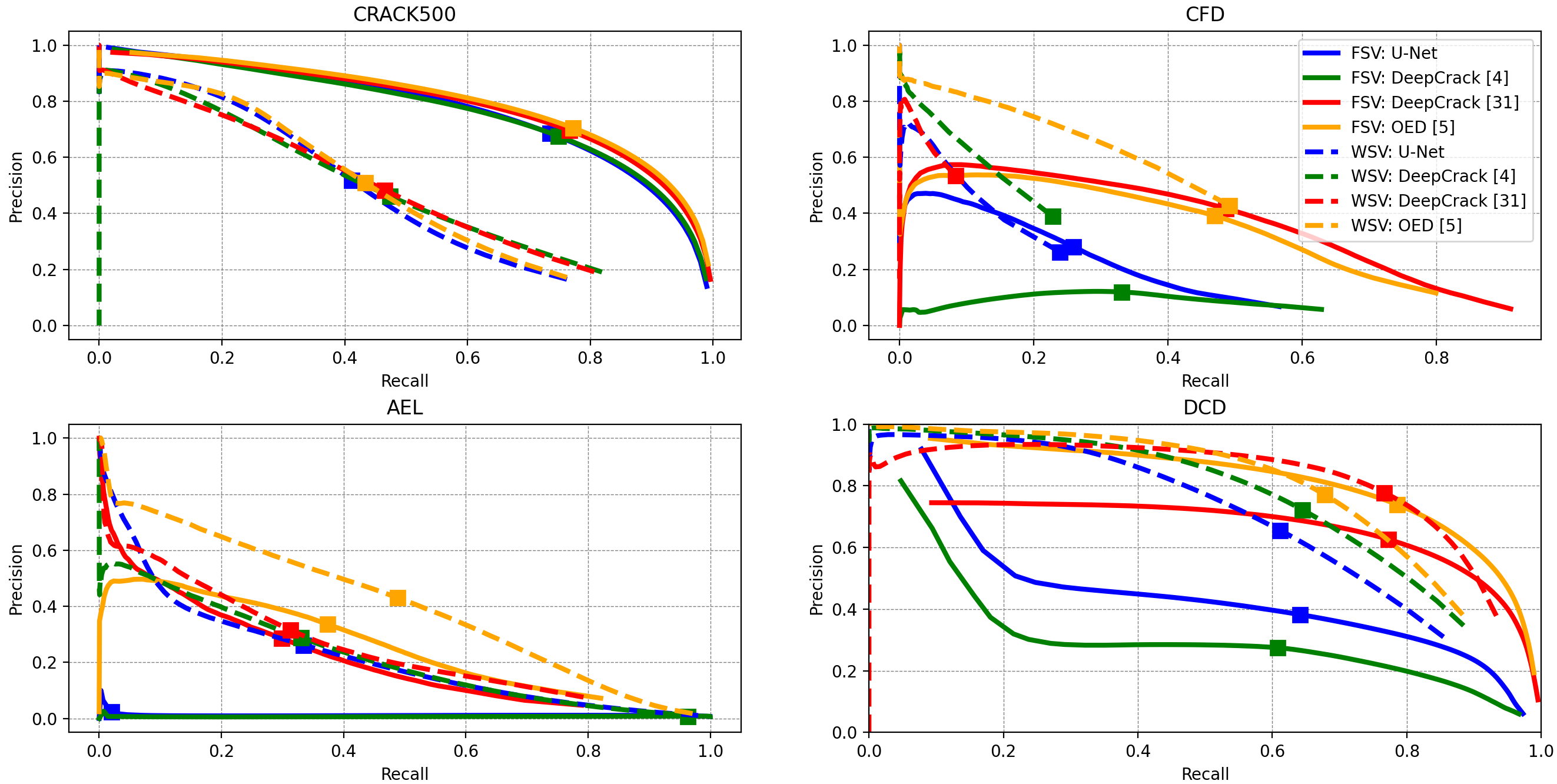}
    \caption{Precision and recall curves of the segmentation results in the four different dataset comparing our weakly-supervised approach (WSV) with training using accurate, fully-supervised, labels (FSV).}
    \label{fig:pr_curves}
\end{figure*}

Example segmentation results that have been generated by using the pseudo labels of our proposed method in comparison with the ones from Dong \textit{et al.} \cite{dong2020PatchBasedWeakly}, as well as fully-supervised learning are shown in \autoref{fig:results}. Additionally, we also show Precision-Recall curves of these methods on the test-datasets in \autoref{fig:pr_curves}.
\subsubsection{CRACK500 Results}
The results on the CRACK500 test set, as shown in \autoref{tab:resultsC500}, show that the fully-supervised methods achieve the highest performance, followed by the segmentation algorithms trained using the CRF based weakly-supervised labels and trailed by the methods trained with the pseudo labels that we proposed.
\subsubsection{CFD Results}
\autoref{tab:resultsCFD} shows the results on the CFD dataset. It can be seen that the segmentation algorithms trained using our proposed, weakly-supervised, pseudo-labels achieve competitive performance in comparison against using the pixel-level accurate labels in the fully-supervised training in all but the DeepCrack \cite{zou2018DeepCrackLearning} segmentation model. Our method is also shown to generally outperform the CRF generated pseudo labels in all but the DeepCrack \cite{zou2018DeepCrackLearning} segmentation model.
\subsubsection{AEL Results}
Results on the AEL dataset are shown in \autoref{tab:resultsAEL}. It can be seen that all segmentation algorithms achieve higher performance than the other methods when trained using our proposed pseudo labels. 
\subsubsection{DCD Results}
The results on the DCD dataset in \autoref{tab:resultsDCD} show that the segmentation results of all methods but the OED \cite{konig2021OptimizedDeep}, results using our weakly-supervised pseudo-labeling approach achieve better performance than the fully-supervised models. Additionally, it also shows that on this dataset, across all models, the CRF labels have achieved the worse segmentation performance.

Summarising those results, it can be seen that our weakly-supervised method achieve close to comparable performance across various crack-segmentation algorithms, in all but the CRACK500 dataset, as if they would have been trained using pixel-level annotations. It is also shown that our approach creates better pseudo-labels as the approach by Dong \textit{et al} \cite{dong2020PatchBasedWeakly} in those datasets, which consequently leads to a better performance on the end-to-end segmentation task in a majority of the datasets. The shortcomings in the performance of our method in comparison against the others on the CRACK500 dataset are further elaborated on in \autoref{sec:crack500just}.

\begin{table}[!tb]
    \centering
    \caption{Results of OIS, ODS and AP on the CRACK500 test set split by their learning approach.}
    \renewcommand{\arraystretch}{0.8}

\begin{tabular}{lllll}\toprule
Label-Type & Method & ODS & OIS & AP \\\midrule

FSV & U-Net & \underline{0.6631} & \underline{0.7148} & \underline{0.7716} \\ 
 & DeepCrack \cite{liu2019DeepCrackDeepa} & \underline{0.6620} & \underline{0.7197} & \underline{0.7670}\\ 
 & DeepCrack \cite{zou2018DeepCrackLearning} & \underline{0.6860} & \underline{0.7303} & \underline{0.7902} \\ 
 & OED \cite{konig2019ConvolutionalNeurala} & \underline{0.6911} & \underline{0.7373}
 & \underline{0.8004} \\ 
 \midrule
WSV & U-Net & 0.5588 & 0.6320 & 0.6517 \\ 
(CRF Labels) & DeepCrack \cite{liu2019DeepCrackDeepa} & 0.5664 & 0.6373 & 0.6513 \\ 
 & DeepCrack \cite{zou2018DeepCrackLearning} & 0.5793 & 0.6444 & 0.6223 \\ 
 & OED \cite{konig2019ConvolutionalNeurala} & 0.5201 & 0.6114 & 0.5402 \\ 
 \midrule
WSV & U-Net & 0.4468 & 0.5443 & 0.4371 \\ 
(Our Labels) & DeepCrack \cite{liu2019DeepCrackDeepa} & 0.4504 & 0.5669 & 0.4546 \\ 
 & DeepCrack \cite{zou2018DeepCrackLearning} & 0.4539 & 0.5602 & 0.4508 \\ 
 & OED \cite{konig2019ConvolutionalNeurala} & 0.4550 & 0.5494 & 0.4472 \\ 

\bottomrule
\end{tabular}

    \label{tab:resultsC500}
\end{table}

\begin{table}[!tb]
    \centering
    \caption{Results of OIS, ODS and AP on the CFD dataset set split by their learning approach.}
    \renewcommand{\arraystretch}{0.8}
\begin{tabular}{lllll}\toprule
Label-Type & Method & ODS & OIS & AP \\\midrule

FSV & U-Net & \underline{0.1846} & \underline{0.2699} & \underline{0.1557} \\ 
 & DeepCrack \cite{liu2019DeepCrackDeepa} & 0.1667 & \underline{0.2435} & 0.0631 \\ 
 & DeepCrack \cite{zou2018DeepCrackLearning} & \underline{0.3940} & \underline{0.4632} & \underline{0.3487} \\ 
 & OED \cite{konig2019ConvolutionalNeurala} & 0.3682 & \underline{0.4541} & 0.3116 \\ 
 \midrule
WSV & U-Net & 0.0909 & 0.1674 & 0.0958 \\ 
(CRF Labels) & DeepCrack \cite{liu2019DeepCrackDeepa} & 0.0856 & 0.1446 & 0.0772 \\ 
 & DeepCrack \cite{zou2018DeepCrackLearning} & 0.2018 & 0.2625 & 0.1479 \\ 
 & OED \cite{konig2019ConvolutionalNeurala} & 0.1042 & 0.1589 & 0.0443 \\ 
 \midrule
WSV & U-Net & 0.1841 & 0.1953 & 0.1218 \\ 
(Our Labels) & DeepCrack \cite{liu2019DeepCrackDeepa} & \underline{0.1874} & \underline{0.1941} & 0.1488 \\ 
 & DeepCrack \cite{zou2018DeepCrackLearning} & 0.0762 & 0.0769 & 0.0693 \\ 
 & OED \cite{konig2019ConvolutionalNeurala} & \underline{0.4022} & 0.4330 & \underline{0.3413} \\ 

\bottomrule
\end{tabular}
    \label{tab:resultsCFD}
\end{table}

\begin{table}[!tb]
    \centering
    \caption{Results of OIS, ODS and AP on the AEL dataset set split by their learning approach.}
    \renewcommand{\arraystretch}{0.8}
\begin{tabular}{lllll}\toprule
Label-Type & Method & ODS & OIS & AP \\\midrule

FSV & U-Net & 0.0478 & 0.0599 & 0.0112 \\ 
 & DeepCrack \cite{liu2019DeepCrackDeepa} & 0.0334 & 0.0390 & 0.0069 \\ 
 & DeepCrack \cite{zou2018DeepCrackLearning} & 0.2025 & 0.2823 & 0.2021 \\ 
 & OED \cite{konig2019ConvolutionalNeurala} & 0.2942 & 0.3519 & 0.2432 \\ 
 \midrule
WSV & U-Net & 0.1555 & 0.2028 & 0.0468 \\ 
(CRF Labels) & DeepCrack \cite{liu2019DeepCrackDeepa} & 0.0683 & 0.1114 & 0.0289 \\ 
 & DeepCrack \cite{zou2018DeepCrackLearning} & 0.1691 & 0.2165 & 0.0899 \\ 
 & OED \cite{konig2019ConvolutionalNeurala} & 0.1016 & 0.1259 & 0.0365 \\ 
 \midrule
WSV & U-Net & \underline{0.2550} & \underline{0.3345} & \underline{0.2163} \\ 
(Our Labels) & DeepCrack \cite{liu2019DeepCrackDeepa} & \underline{0.2766} & \underline{0.3967} & \underline{0.2134} \\ 
 & DeepCrack \cite{zou2018DeepCrackLearning} &\underline{0.3830} & \underline{0.4604} & \underline{0.2391} \\ 
 & OED \cite{konig2019ConvolutionalNeurala} & \underline{0.4461} & \underline{0.5022} & \underline{0.4024}\\ 

\bottomrule
\end{tabular}
    \label{tab:resultsAEL}
\end{table}

\begin{table}[!tb]
    \centering
    \caption{Results of OIS, ODS and AP on the DCD dataset set split by their learning approach.}
    \renewcommand{\arraystretch}{0.8}
\begin{tabular}{lllll}\toprule
Label-Type & Method & ODS & OIS & AP \\\midrule

FSV & U-Net & 0.5886 & 0.6475 & 0.4478 \\ 
 & DeepCrack \cite{liu2019DeepCrackDeepa} & 0.4643 & 0.5497 & 0.3095 \\ 
 & DeepCrack \cite{zou2018DeepCrackLearning} & 0.6521 & 0.6954 & 0.6653 \\ 
 & OED \cite{konig2019ConvolutionalNeurala} & \underline{0.7055} & 0.7474 & \underline{0.8148} \\ 
 \midrule
WSV & U-Net & 0.4417 & 0.5432 & 0.5655 \\ 
(CRF Labels) & DeepCrack \cite{liu2019DeepCrackDeepa} & 0.3705 & 0.4395 & 0.4431 \\ 
 & DeepCrack \cite{zou2018DeepCrackLearning} & 0.4457 & 0.4961 & 0.4889 \\ 
 & OED \cite{konig2019ConvolutionalNeurala} & 0.2616 & 0.3271 & 0.0942 \\ 
 \midrule
WSV & U-Net & \underline{0.6199} & \underline{0.7120} & \underline{0.6595} \\ 
(Our Labels) & DeepCrack \cite{liu2019DeepCrackDeepa} & \underline{0.6597} & \underline{0.7319} & \underline{0.7228} \\ 
 & DeepCrack \cite{zou2018DeepCrackLearning} & \underline{0.6936} & \underline{0.7530} & \underline{0.7961} \\ 
 & OED \cite{konig2019ConvolutionalNeurala} & 0.6843 & \underline{0.7605} & 0.7596 \\ 

\bottomrule
\end{tabular}
    \label{tab:resultsDCD}
\end{table}

\subsection{Ablation Studies}
To show how the components of our method contribute to its performance we have performed ablation studies. In \autoref{tab:abl_results} we report results that leave out either one of the morphology or bilateral filtering steps and the method that only performs two-class Otsu thresholding on the patches which yields only a single threshold. This is shown on the CRACK500 training dataset by comparing our proposed pseudo labels with the actual labels of this dataset.
As it can be seen, both morphology and the bilateral filtering operation add an increase in the pseudo label quality. However, the inclusion of three-class Otsu thresholding in comparison with two-class thresholding yields a slightly lower performance. We attribute this due to the intensity distribution, as in the CRACK500 dataset cracks seem to have a much darker profile than the surrounding backgrounds. However, as this can differ from dataset to dataset, we opt to, and encourage the use of three-class thresholding as we found that empirically it gives better results on datasets where the cracks are only slightly darker than their background. The very slight drop in performance on datasets with darker cracks is made up in a higher label quality on datasets where the cracks are brighter,  as less background pixels are labelled as cracks. 

\begin{table}[!tb]
    \centering
    \caption{Ablation experiments by comparing the pseudo-labels with the original labels on the CRACK500 train set}
    \renewcommand{\arraystretch}{0.8}
    \begin{tabular}{llll}\toprule
        Ablation & ODS & OIS & AP  \\\midrule
        Our-Labels, w/o Bilateral   & 0.4691 & 0.5034 &  0.3539   \\
        Our-Labels, w/o Morphology  & 0.4436  & 0.4436  &  0.2478 \\
        Our-Labels, with 2-Class Otsu  & \underline{0.5375}  & \underline{0.6128}  &  \underline{0.5041} \\\midrule
        Our-Labels & 0.5359 & 0.5839 & 0.4371 \\

    \bottomrule
    \end{tabular}

    \label{tab:abl_results}
\end{table}

\subsection{Bells and Whistles: Semi Supervised}
We also show that our approach can achieve a slightly better performance when incorporating further unlabelled data into the training-set corpus. For this we use the CRACK500 test-set and run our pseudo-labelling approach on it. Then, we incorporate these further labels into the training process and train the segmentation algorithms as described in \autoref{sec:e2e_seg}. \autoref{tab:bells_whistles} shows the performance when additional data is added in comparison to only training on the data that has classification labels. 
It can be seen that whilst on CRACK500 and DCD comparable results are achieved, the added data has benefited the results on CFD and AEL more, adding a performance increase 1.98\%, 2.27\% and 2.26\% for ODS, OIS and AP on the CFD, and 2.37\%, 1.85\% and 5.65\% on AEL respectively.
This shows that the addition of further, unlabelled data generally either keeps the performance similar or increases it on algorithms that have been trained using our weakly-supervised approached. 

\begin{table}[!tb]
    \centering
    \caption{Results of OIS, ODS and AP by incorporating pseudo labels of the test set into the training the OED \cite{konig2021OptimizedDeep} method, making this approach semi-supervised.}
    \renewcommand{\arraystretch}{0.8}
    \begin{tabular}{lllll}\toprule
    Dataset & Pseudo-Labels & ODS & OIS & AP \\
    \midrule
    Crack500 & Train &  \underline{0.4550} & 0.5494 & 0.4472 \\ 
        & Train + Test   & 0.4562 & \underline{0.5567} & \underline{0.4507} \\
    \midrule
    CFD & Train &  0.4022 & 0.4330 & 0.3413 \\ 
        & Train + Test  & \underline{0.4220} & \underline{0.4557} & \underline{0.3639}\\
    \midrule
    AEL & Train &  0.4461 & 0.5022 & 0.4024\\ 
        & Train + Test & \underline{0.4698} & \underline{0.5207} & \underline{0.4589}\\
    \midrule
    DCD & Train &  \underline{0.6843} & \underline{0.7605} & \underline{0.7596} \\ 
        & Train + Test  & 0.6781 & 0.7582 & 0.7442 \\
    \bottomrule

    \end{tabular}
    \label{tab:bells_whistles}
\end{table}

\begin{figure}[!t]
\centering
\includegraphics[width=\linewidth]{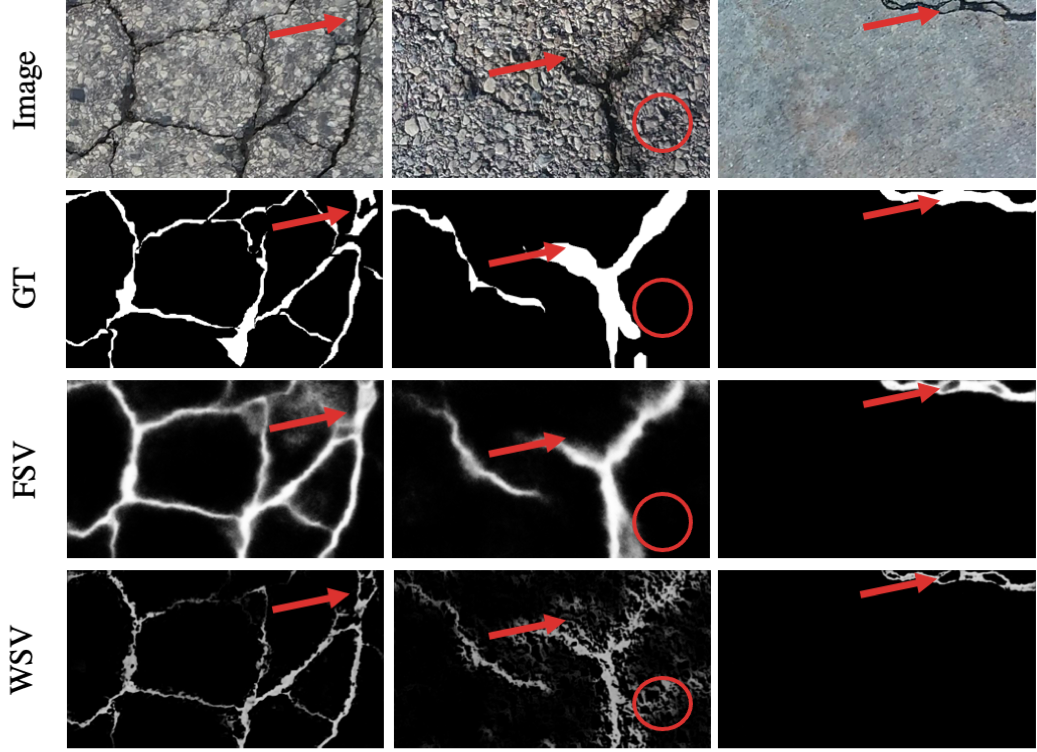}
\caption{Slight inaccuracies in the GT (arrows) as well as issues of our method when the background is highly textured (circle).}
\label{fig:incorrect}
\end{figure}

\subsection{Discussion}
\subsubsection{Performance Drop on CRACK500}
\label{sec:crack500just}
In comparison to the fully-supervised method or using CRF for pseudo labels, our method performs worse on CRACK500. We hypothesize this is due to the lower annotation quality on this dataset in comparison to DCD, AEL and CFD and the amount of texture in the background through pebbles or similar obstructions. \autoref{fig:incorrect} shows some example images of the test set, where the labels are either slightly inaccurate or have a highly textured backgrounds and predictions of our method using OED utilizing our WSV and the standard FSV labels. As it can be seen, on small inclusions such as stones, whilst being within a crack, are picked up by our method and labelled as not containing a crack, whereas the FSV method and the original GT have this area as belonging to a crack.  To validate our hypothesis we show results of the OED method which has been trained on a combined dataset with the training set of DCD, 71 images from CFD and 38 images of AEL using the same hyperparameters as stated previously, and test it on the remaining images. The results are shown in \autoref{tab:resultsDCDCFDAEL}. It can be seen that when the training and testing labels are more accurate, our method outperforms the CRF approach, showing that it generates better predictions \textit{without} the need to perform a grid search to find the best parameters and is able to achieve competitive results against the fully-supervised method.
\subsubsection{Other Datasets}
In the CFD, AEL and DCD datasets the tabular results and precision recall curves in \autoref{fig:pr_curves} show at least competitive, and even slight increases in performance when using our weakly-supervised method in comparison with the fully-supervised approach. 
Our method is able to accurately segment large regions of cracks, whist still keeping fine boundary details in contrast to the fully-supervised and weakly-supervised approach using CRF, which appear to have blurred segmentation maps at crack boundaries. 
It is to note, that when using our pseudo-labelling approach on very large images, the class activation mapping seems to only focus on the very descriptive features which can hinder localization performance. We therefore recommend that when larger images are used, they should be split into the size on which the classifier was trained, before applying our weakly-supervised segmentation method to achieve the highest possible performance. The results in \autoref{tab:resultsDCDCFDAEL} show that when using datasets with higher accuracy in the annotations and thinner cracks, our method outperforms the CRF method by Dong \textit{et al.} \cite{dong2020PatchBasedWeakly} in the ODS and OIS metrics and only performs a little worse in two of the tested segmentation algorithms in the AP metric. When observing the results from the OED method, the best performing segmentation algorithm on the combined DCD, CFD and AEL datasets, our WSV approach achieves only a 3.48\% lower ODS, 3.14\% lower OIS and 7.86\% lower AP in comparison to the fully-supervised method. However, this slightly lower performance might be negligible if the labelling time for the training data is reduced significantly.
Even though a drawback of our pseudo-labelling method is the initial labelling speed to create the pseudo-labels, which is exponentially slower than the CRF method, it makes up for it in the increase in performance that can be achieved once the pseudo labels have been created. Another advantage in comparison to the method by Dong \textit{et al.} \cite{dong2020PatchBasedWeakly} is also that no grid-search is necessary to obtain the best performing hyperparameters.

\begin{table}[!tb]
    \centering
    \caption{Results of OIS, ODS and AP when trained and tested on a combination of the DCD, CFD and AEL datasets}
    \renewcommand{\arraystretch}{0.8}
\begin{tabular}{lllll}\toprule
Label-Type & Method & ODS & OIS & AP \\\midrule

FSV & U-Net & \underline{0.7347} & \underline{0.7910} & \underline{0.8861} \\ 
 & DeepCrack \cite{liu2019DeepCrackDeepa} & \underline{0.7347} & \underline{0.7948} & \underline{0.8696} \\ 
 & DeepCrack \cite{zou2018DeepCrackLearning} & \underline{0.5744} & \underline{0.6275} & \underline{0.6511} \\ 
 & OED \cite{konig2019ConvolutionalNeurala} & \underline{0.7581} & \underline{0.8119} & \underline{0.8965} \\ 
 \midrule
WSV & U-Net & 0.6523 & 0.7345 & 0.8175 \\ 
(CRF Labels) & DeepCrack \cite{liu2019DeepCrackDeepa} & 0.6286 & 0.7242 & 0.8227\\ 
 & DeepCrack \cite{zou2018DeepCrackLearning} & 0.4178 & 0.4914 & 0.4370\\ 
 & OED \cite{konig2019ConvolutionalNeurala} & 0.6641 & 0.7321 & 0.8035\\ 
 \midrule
WSV & U-Net & 0.6700 & 0.7371 & 0.7511 \\ 
(Our Labels) & DeepCrack \cite{liu2019DeepCrackDeepa} & 0.7036 & 0.7674 & 0.8068 \\ 
 & DeepCrack \cite{zou2018DeepCrackLearning} & 0.5367 & 0.5774 & 0.5879 \\ 
 & OED \cite{konig2019ConvolutionalNeurala} & 0.7233 & 0.7805 & 0.8179 \\ 

\bottomrule
\end{tabular}

    \label{tab:resultsDCDCFDAEL}
\end{table}

\section{Conclusion}
In this paper we have proposed a method that expands pretrained crack classification CNNs and leverages them to create a pseudo-label output of surface cracks. Those created labels are then incorporated into end-to-end segmentation network training. In comparison with other prevalent methods for weakly-supervised segmentation, we do not use a conditional random field, but rather utilise a novel patch-threshold segmentation approach merged with a rough localization map generated by the classifier.
We use this pseudo labeling approach to create pixel-level segmentation labels of images that previously only had a classification annotation and train multiple recent crack-segmentation algorithms with this.
The use of our method for weakly-supervised segmentation is shown to achieve better results in four out of five popular crack datasets, outperforming or achieving comparable performance to the fully-supervised approach. We also elaborate on the reason for the performance drop on the dataset where it performs worse, attributing it to the image and label quality in this particular set of data and provide results showing that if the quality is increased, or other datasets are used, our method can achieve results coming close to fully-supervised methods, only requiring an exponentially smaller labelling effort in the initial surface crack images.
For future work we would aim to study other adaptive thresholding techniques that can be utilized for pseudo-labeling as the limitations of our method lies within very large cracks that take up large portions of images.

\IEEEpeerreviewmaketitle

\appendices

\ifCLASSOPTIONcaptionsoff
  \newpage
\fi

\bibliographystyle{IEEEtran}
\bibliography{ref}


\end{document}